\documentclass[conference,table,xcdraw]{IEEEtran}
\IEEEoverridecommandlockouts
\usepackage{cite}
\usepackage{amsmath,amssymb,amsfonts}
\usepackage[nolist,nohyperlinks]{acronym}
\usepackage{float} % Add this in your preamble if not already included

\usepackage{algorithmic}
\usepackage{graphicx}
\usepackage{tabularx}
\usepackage{textcomp}
\usepackage{xcolor}
\def\BibTeX{{\rm B\kern-.05em{\sc i\kern-.025em b}\kern-.08em
    T\kern-.1667em\lower.7ex\hbox{E}\kern-.125emX}}
\usepackage{float}
\usepackage{amsfonts}
\usepackage{subcaption}
\usepackage{caption}
\usepackage{booktabs}
{}
{}
{}
\usepackage{tikz}
\usetikzlibrary{shapes, arrows}
\usepackage{pdfpages}
\usepackage{pgfplots}
\usetikzlibrary{arrows,automata}
\usepackage[utf8]{inputenc}
\usepackage{multirow}
\usepackage{blindtext}
\usepackage{lipsum}% http://ctan.org/pkg/lipsum
\usepackage{pgfplots}
\usepackage{mathtools}

\usepackage{balance}
\usepackage{comment}
\usepackage{amsmath}
\usepackage{graphics}
\usepackage{pgfplots,tikz}
\usepackage{xcolor}
\usepackage{pagecolor}% http://ctan.org/pkg/{pagecolor,lipsum}
\usetikzlibrary{decorations, decorations.text,backgrounds}
\colorlet{NextBlue}{red!25!green!50!blue!75}
\usepackage{hyperref}
\usepackage{verbatim}
\usetikzlibrary{shapes}
\tikzstyle{line} = [draw,-latex']
\usepackage{array}
\usepackage{geometry}
\usepackage{amsmath}
\usepackage{hyperref}
\usepackage{amsfonts}
\usepackage{algorithm}
\usepackage{algorithmic}
\begin{document}

\title{ChatGPT or A Silent Everywhere Helper: A Survey of Large Language Models \\
%LoS ansd NLoS identification in MIMO-based UAVs USING Deep Learning*\\

\thanks{Identify applicable funding agency here. If none, delete this.}
}

\author{\IEEEauthorblockN{1\textsuperscript{st} Azim Akhtarshenas}
\IEEEauthorblockA{\textit{dept. of Telecommunication Engineering} \\
\textit{Universitat Polit\`ecnica de Val\`encia}\\
Valencia, Spain\\
aakhtar@doctor.upv.es}
\and
\IEEEauthorblockN{2\textsuperscript{nd} Afshin Dini}
\IEEEauthorblockA{\textit{dept. of Computer Science} \\
\textit{Tampere University}\\
Tampere, Finland \\
afshin.dini@tuni.fi}
\and
\IEEEauthorblockN{3\textsuperscript{nd} Navid Ayoobi}
\IEEEauthorblockA{\textit{dept. of Computer Science} \\
\textit{University of Houston}\\
Houston, USA \\
nayoobi@cougarnet.uh.edu}}
%\and
%\IEEEauthorblockN{3\textsuperscript{rd} David Lopez-Perez}
%\IEEEauthorblockA{\textit{dept. of Telecommunication Engineering} \\
%\textit{Universitat Polit\`ecnica de Val\`encia}\\
%Valencia, Spain \\
%D.lopez@iteam.upv.es}
%\and
%\IEEEauthorblockN{4\textsuperscript{th} Ramin Toosi}
%\IEEEauthorblockA{\textit{dept. of Electrical and Computer Engineering} \\
%\textit{University of Tehran}\\
%Tehran, Iran\\
%r.toosi@ut.ac.ir}
%\and
%\IEEEauthorblockN{5\textsuperscript{th} Matin Amoozadeh}
%\IEEEauthorblockA{\textit{dept. of Computer Science} \\
%\textit{University of Houston}\\
%Houston, USA \\
%mamoozad@cougarnet.uh.edu}
%}
%\and
%\IEEEauthorblockN{2\textsuperscript{nd} Navid Ayoobi}
%\IEEEauthorblockA{\textit{dept. of Computer Science} \\
%\textit{University of Houston}\\
%Houston, USA \\
%nayoobi@cougarnet.uh.edu}
%\and
%\IEEEauthorblockN{3\textsuperscript{rd} David Lopez-Perez}
%\IEEEauthorblockA{\textit{dept. of Telecommunication Engineering} \\
%\textit{Universitat Polit\`ecnica de Val\`encia}\\
%Valencia, Spain \\
%D.lopez@iteam.upv.es}
%\and
%\IEEEauthorblockN{4\textsuperscript{th} Ramin Toosi}
%\IEEEauthorblockA{\textit{dept. of Electrical and Computer Engineering} \\
%\textit{University of Tehran}\\
%Tehran, Iran\\
%r.toosi@ut.ac.ir}
%\and
%\IEEEauthorblockN{5\textsuperscript{th} Matin Amoozadeh}
%\IEEEauthorblockA{\textit{dept. of Computer Science} \\
%\textit{University of Houston}\\
%Houston, USA \\
%mamoozad@cougarnet.uh.edu}
%}
\maketitle
\begin{abstract}
~\acp{LLM} have revolutionized natural language processing~\ac{NLP}, with~\ac{ChatGPT} standing out as a notable example due to its advanced capabilities and widespread applications. This survey provides a comprehensive analysis of~\ac{ChatGPT}, exploring its architecture, training processes, and functionalities. We examine its integration into various domains across industries such as customer service, education, healthcare, and entertainment. A comparative analysis with other~\acp{LLM} highlights~\ac{ChatGPT}'s unique features and performance metrics. Regarding benchmarks, the paper examines ChatGPT’s comparative performance against other~\acp{LLM} and discusses potential risks such as misinformation, bias, and data privacy concerns. 
Additionally, we offer a number of figures and tables that outline the backdrop of the discussion, the main ideas of the article, the numerous~\ac{LLM} models, a thorough list of datasets used for pre-training, fine-tuning, and evaluation, as well as particular~\ac{LLM} applications with pertinent references. Finally, we identify future research directions and technological advancements, underscoring the evolving landscape of~\acp{LLM} and their profound impact on artificial intelligence \ac{AI} and society.
\end{abstract}
\section{Introduction}\label{Intro}
\begin{figure}
    \centering
    \begin{tikzpicture}[scale=0.8] % Scale the whole figure to 80% of its original size
        \begin{axis}[
            xlabel={Year},
            ylabel={Number of Papers},
            xtick={2020, 2021, 2022, 2023, 2024},
            grid=major,
            ymin=0, ymax=1600,
            ytick={0, 250, 500, 750, 1000, 1250, 1500},
            mark options={solid},
            legend pos=north west
        ]
        \addplot[
            color=blue,
            mark=square,
            ]
            coordinates {
            (2020,50)
            (2021,200)
            (2022,500)
            (2023,1000)
            (2024,1500)
            };
        \end{axis}
    \end{tikzpicture}
    \caption{Number of Published Papers on ChatGPT and LLM (2020-2024)}
    \label{publsihed_paper}
\end{figure}
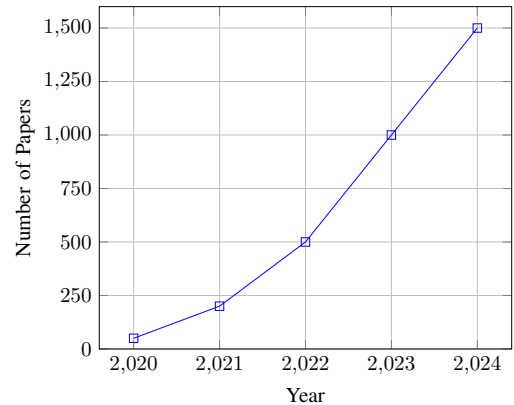

\textbf{\textit{~\ac{LLM} can significantly enhance a robot's ability to understand and respond to human emotions.  By analyzing voice and text, ~\acp{LLM} can recognize emotional and contextual clues, allowing robots to provide empathetic and contextually appropriate responses.  This ability could lead to more effective and organic human-robot interactions.  Although ~\acp{LLM} can aid in the comprehension of emotions, more knowledge about nonverbal indicators such as facial expressions and body language, together with advanced models, is required to completely understand and correctly interpret complex human emotions.}}\\
\begin{comment}
{\color{red}Overview of~\acp{LLM}: Briefly introduce Large Language Models, their evolution, and significance.\\
Focus on ChatGPT: Introduce ChatGPT as a case study or focal point. \\
}
\end{comment}
The invention of language stands as a pivotal milestone in human history, fundamentally transforming the fabric of society and bridging the gap between individuals. Due to its complexity and endless rules, this groundbreaking development revolutionized communication, enabling the transmission of encoded ideas, enhancing cultural evolution, and facilitating cooperative endeavors \cite{zhao2023survey}. Another revolution is currently underway, where machines and devices are being designed to understand, evaluate, and process languages using~\acp{LLM} \cite{raiaan2024review}. These advancements allow them to predict various words or even sentences that have been hidden or lost for different reasons using~\ac{NLP} \cite{bimagambetova2023evaluating}. 
Due to the revolutions above and numerous other advancements not detailed here,~\acp{LLM} have garnered significant attention from both the scientific community and industry professionals. Researchers are increasingly exploring the potential of~\acp{LLM} to push the boundaries of artificial intelligence, while industry leaders are keenly interested in their applications across various sectors, including healthcare, finance, and customer service. This growing interest is driven by the transformative capabilities of~\acp{LLM} to process and generate human-like text, enabling innovative solutions and efficiencies that were previously unattainable. The number of published papers in this area is a clear indicator of the growing research interest and activity (See Fig. \ref{publsihed_paper}).

\subsection{Previous surveys on~\ac{LLM} and their gaps}
In this section, we aim to explore the previous surveys about~\ac{LLM} and~\ac{ChatGPT} and investigate their strengths and weaknesses. Numerous studies and surveys have been conducted to evaluate the performance, utility, and impact of these advanced~\ac{AI} systems across different domains. These works provide critical insights into how~\acp{LLM} and~\ac{ChatGPT} are being utilized, their efficacy in various applications, and the challenges they present. By analyzing these surveys, we can better understand the current state of these technologies, identify common themes in user experiences, and highlight areas where improvements are necessary. This exploration will also shed light on the broader implications of integrating such AI systems into everyday use, including ethical considerations, user satisfaction, and technical limitations.\\

In this regard, the authors in \cite{hadi2023large} explore~\acp{LLM}, detailing their origins, architecture, training methods, applications, and challenges. It begins with generative~\ac{AI} concepts and the design of~\ac{GPT}, then traces the historical development and training techniques of~\acp{LLM}. Applications in fields like medicine, education, and finance are discussed, along with their role in~\ac{AI}'s future and scientific breakthroughs. Challenges such as ethical issues, biases, interpretability, and computational demands are examined. Methods to enhance~\acp{LLM} robustness and control are highlighted. Finally, the analysis outlines future research directions to improve the reliability and utility of~\ac{LLM} technology.
The authors in \cite{zhao2023survey, naveed2023comprehensive} commence with an examination of early pre-trained language models such as~\ac{BERT}. Subsequently, they delve into three prominent~\ac{LLM} families, including~\ac{GPT},~\ac{LLaMA}, and~\ac{PaLM}, alongside other noteworthy LLM variants. The exploration extends to diverse methodologies for constructing, enhancing, and harnessing~\acp{LLM}. Furthermore, we scrutinize prevalent~\ac{LLM} datasets and benchmarks, juxtaposing the efficacy of various prominent models across public benchmarks. \\

The authors in \cite{chang2024survey} aim to address three main questions "what to evaluate, where to evaluate, and how to evaluate." To this end, an overview is provided of evaluation tasks for~\acp{LLM} and~\ac{ChatGPT}, encompassing general natural language processing, reasoning, medical applications, ethics, education, natural and social sciences, and agent-based applications. The evaluation methods and benchmarks, which are crucial for assessing the performance of~\acp{LLM}, are examined to address the 'where' and 'how' questions. Additionally, the success and failure cases of~\acp{LLM} in different tasks are summarized.\\
%5
The survey \cite{raiaan2024review} provides a comprehensive overview of~\acp{LLM}, covering their history, architectures, transformer mechanisms, resources, training methods, applications, impacts, and challenges. It starts by explaining the fundamental concepts and traditional training pipeline of~\acp{LLM}. Next, it reviews existing research, tracing the evolution of~\acp{LLM} and detailing the architecture of transformers, the various resources employed, and the diverse training techniques used. The datasets utilized in these studies are also highlighted. The article then explores the wide-ranging applications of~\acp{LLM} across fields such as biomedical and healthcare, education, social sectors, business, and agriculture. Additionally, it discusses the societal impact of~\acp{LLM}, their role in the future shape of AI, and their potential to address real-world problems. \\
Research attention has long been directed towards autonomous agents in both academic and industry circles.

%6
In \cite{wang2024survey}, existing research in the field of~\acp{LLM}-based autonomous agents is systematically summarized. These studies are presented and reviewed from three aspects: the construction, application, and evaluation of the agents. For each of these aspects, a detailed taxonomy is provided to draw connections among the existing research, summarizing the major techniques and their development histories. \\

%7
~\acp{LLM} play important role in understanding and producing human-like text but face significant computational challenges during training due to their extensive parameters. This issue is compounded by the need for frequent updates to keep the models current with evolving information. Applications often require continual adjustments post-training to correct deficiencies and undesirable behaviors. Recently, there has been growing attention to useful, lightweight methods for on-the-fly approach improvement, leading to advancements in knowledge editing techniques. These methods aim to modify~\acp{LLM} behaviors within specific domains while maintaining overall performance. The authors in \cite{zhang2024comprehensive} define the knowledge editing problem, review recent approaches and techniques, introduce the KnowEdit benchmark for empirical evaluation, and provide an analysis of knowledge location within~\acp{LLM}. To support future research, the open-source framework EasyEdit1 has been released, enabling flexible implementation of knowledge editing. Finally, the paper discusses the broad implications and potential applications of these techniques. \\

Despite significant advancements in tasks like translation, summarization, information retrieval, and language generation, which are attracting growing interest in the CHI community, researchers hold diverse and often controversial views on the efficacy, ethics, and cognitive abilities of~\acp{LLM}. In light of these discussions, there is a limited understanding of how people perceive~\acp{LLM} integrated into everyday tools, particularly regarding experiences with bias, safety, social norms, and stereotypes. In \cite{wang2023people}, the authors perform a systematic review to gather empirical insights on public perceptions of~\acp{LLM}. From an initial pool of 231 papers, they conduct a full-text review of 15 studies that engaged human evaluators to assess their interactions with~\acp{LLM}. They present the biases and related concepts examined in these studies, categorize the four main application areas of~\acp{LLM}, and detail the evaluators' perceptions of~\acp{LLM} performance, including conflicting views, biases, and advantages. Additionally, the authors identify factors influencing these perceptions and outline concerns related to~\ac{LLM} applications.\\

~\acp{LLM} have captured the interest of scholars and researchers in security domains, revealing vulnerabilities and showcasing their capabilities in tasks related to security. For example, the authors in \cite{yao2024survey} delves into the intersection of ~\acp{LLM} with security and privacy, examining their positive impacts, associated risks and threats, and inherent vulnerabilities. Through an extensive review of literature, the paper classifies findings into 'The Good' (beneficial ~\ac{LLM} applications), 'The Bad' (potentially offensive applications), and 'The Ugly' (~\ac{LLM} vulnerabilities and defenses). Noteworthy discoveries include ~\acp{LLM}' ability to enhance code security and data privacy, surpassing traditional methods. However, their human-like reasoning capabilities also render them susceptible to various attacks, particularly at the user level.\\
The authors in \cite{cui2024survey} present the foundational context of~\acp{MLLM}, elucidating their evolution in conjunction with multimodal models employing~\acp{LLM}, and tracing the historical progression of~\ac{AD}. Subsequently, they provide an overview of current~\ac{MLLM} tools contributed to driving, transportation, and mapping systems, along with available datasets and benchmarking methodologies. Furthermore, they briefly explore the contributions presented in The 1st WACV Workshop on~\ac{LLVM-AD}, marking the inaugural workshop dedicated to exploring ~\acp{LLM} applications within~\ac{AD} contexts.
\begin{figure*}[h!]
    \centering
    \includegraphics[width=.8\textwidth]{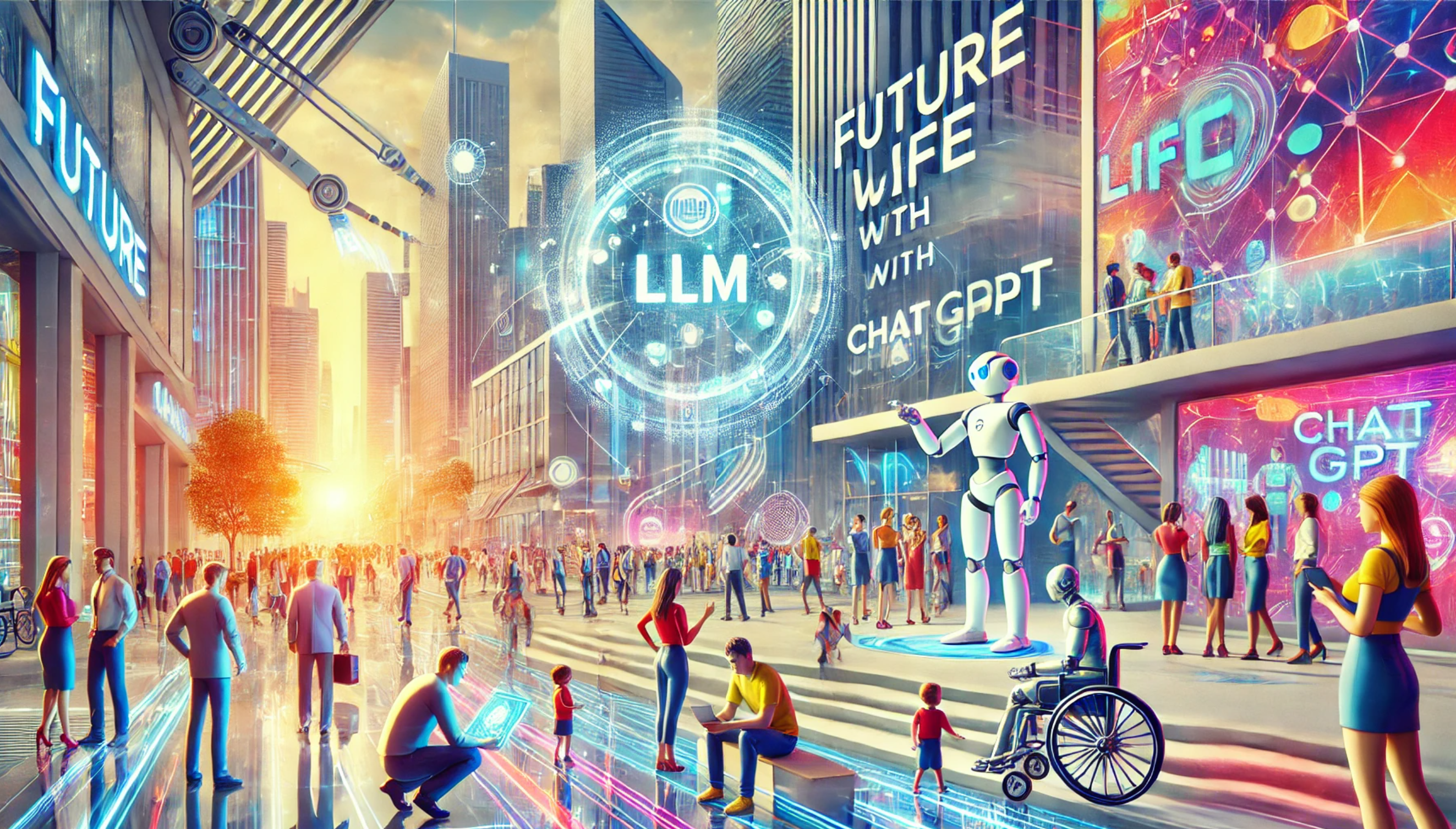}
    \caption{Overview of~\ac{LLM} usage in the future life}
    \label{main_fig}
\end{figure*}
\subsection{Our Contributions, Purpose, and Scope}
Previous surveys on~\acp{LLM} have made significant contributions to the field by providing valuable insights into their capabilities and applications across various tasks. These studies have highlighted the breakthroughs achieved by~\acp{LLM} in translation, summarization, information retrieval, and language generation. However, they have notable weaknesses, such as not thoroughly listing resources, datasets, and models, and failing to explore the crucial relationship between~\acp{LLM} and~\ac{ChatGPT}. These gaps make it challenging for researchers to access vital information and understand how advancements in~\acp{LLM} can enhance~\ac{ChatGPT}. Our paper addresses these shortcomings by offering comprehensive resource documentation, an in-depth analysis of~\ac{ChatGPT}, and a broad overview of~\ac{MLLM} applications, particularly in~\ac{AD}. Additionally, we summarize key insights from The 1st WACV Workshop on~\ac{LLVM-AD}, the first of its kind focusing on~\acp{LLM} in this context. We also discuss future directions for~\ac{LLM} research, emphasizing potential improvements that can significantly impact~\ac{ChatGPT} and similar models. The main contributions of our papers are listed as follows:
\begin{itemize}
\item Comprehensive documentation of resources, datasets, and models relevant to~\ac{LLM} research.
\item Thorough discussion and analysis of~\ac{ChatGPT}, including its architecture and performance.
\item A broad overview of the development and application of MLLMs in various domains.
\item Summary of key insights from The 1st WACV Workshop on~\ac{LLVM-AD}.
\item Identification and discussion of future research directions and potential enhancements for~\ac{ChatGPT}.
\end{itemize}
\subsection{Who benefits?}
\subsubsection{Researchers and Academics}
\begin{itemize}
\item Those studying~\ac{NLP},~\ac{ML}, and~\ac{AI}.
\item Researchers are looking for a comprehensive understanding of the resources, datasets, and models related to~\acp{LLM}.
\item Scholars investigating the ethical, social, and technical implications of~\acp{LLM} and~\ac{ChatGPT}.
\end{itemize}
\subsubsection{Developers and Engineers}
\begin{itemize}
\item Developers are building applications that integrate~\acp{LLM} and~\ac{ChatGPT}.
\item Engineers are focusing on improving the performance and capabilities of~\ac{AI}-driven tools and services.
\end{itemize}
\subsubsection{Data Scientists}
\begin{itemize}
\item Professionals analyze data to improve~\ac{LLM} models and their applications.
\item Those working on data-driven approaches to enhance the accuracy and efficiency of~\ac{ChatGPT}.
\end{itemize}
\subsubsection{Tech Industry Professionals}
\begin{itemize}
\item Product managers and business leaders seeking to incorporate~\acp{LLM} and~\ac{ChatGPT} into their products.
\item Startups and tech companies exploring innovative applications of~\acp{LLM}.
\end{itemize}
\subsubsection{Educational Institutions}
\begin{itemize}
\item Instructors and curriculum developers create educational materials on~\ac{AI},~\ac{ML}, and~\ac{NLP}.
\item Students are learning about the latest advancements and applications of~\acp{LLM} and~\ac{ChatGPT}.
\end{itemize}
\subsubsection{Policy Makers and Ethicists}
\begin{itemize}
\item Individuals concerned with the ethical implications and regulatory aspects of~\ac{AI} technologies.
\item Policymakers are drafting guidelines and frameworks for the safe and ethical use of~\acp{LLM}.
\end{itemize}
\subsubsection{General Public and End-users}
\begin{itemize}
\item Users interested in understanding how~\acp{LLM}, particularly~\ac{ChatGPT}, work and their potential impact.
\item Individuals looking for insights into the future directions and applications of~\ac{AI} in everyday tools.
\end{itemize}
\subsubsection{Business and Marketing Professionals}
\begin{itemize}
\item Marketing teams leverage~\acp{LLM} for content generation, customer interaction, and data analysis.
\item Business analysts assess the market potential and business models involving~\ac{LLM} technologies.
\end{itemize}
\subsubsection{Healthcare Professionals}
Researchers and practitioners explore the application of~\acp{LLM} in medical data analysis, patient interaction, and personalized medicine.
\subsubsection{Legal Professionals}
Lawyers and legal scholars examine the implications of  ~\acp{LLM}-generated content on intellectual property, privacy, and liability issues.
It is clear from Figure~\ref{main_fig} that~\ac{ChatGPT} and~\acp{LLM} will be crucial in determining the future, as their broad use will affect many facets of day-to-day living.

\section{Background}\label{background}
A~\ac{LM} is a computational model which is designed to understand and generate text or speech in various human languages~\cite{devlin2018bert,chang2024survey}. Statistical models~\cite{bellegarda2004statistical} such as N-gram~\cite{pauls2011faster} and Hidden Markov models~\cite{chiu2020scaling}, which tries to estimate the probabilities of word sequences on a context, are the firstly developed language models while further traditional machine learning approaches such as~\ac{SVM}~\cite{yang2006improved} and Naive Bayes~\cite{peng2004augmenting} try to improve the performance of these models.
The most recent and common language models are developed based on neural network approaches as~\ac{RNN}~\cite{xiao2020research} and~\ac{LSTM}~\cite{sundermeyer2012lstm} techniques which also forms the basis of Large Language Models. Although~\acp{LM} can be fit into some applications, they have difficulties to predicts the complex linguistic structures, long dependencies of words, and the rare unseen words~\cite{chang2024survey}. 
To deal with these challenges,~\acp{LLM}~\cite{kasneci2023chatgpt,naveed2023comprehensive,zhao2023survey} as the complex forms of language models with huge number of trainable parameters, are trained on large datasets of texts to give a better representation of the patterns and structures of a language.~\ac{ELMo}~\cite{peters-etal-2018-deep} developed by Allen Institute for Artificial Intelligence Natural Language Processing based on the contextual word embedding,~\ac{GPT}~\cite{radford2018improving}, and~\ac{BERT}~\cite{Devlin2019BERTPO}, developed by OpenAI and Google~\ac{AI} based on the transformers~\cite{NIPS2017_3f5ee243}, are the first~\acp{LLM} introduced in 2018. In 2019,~\ac{GPT}-2~\cite{Radford2019LanguageMA} with a huge amount of trainable parameters improved the text generation capability of~\ac{GPT} while new models like ~\ac{XLNet}~\cite{NEURIPS2019_dc6a7e65} and ~\ac{RoBERTa}~\cite{Liu2019RoBERTaAR}, as optimized versions of BERT, and T5~\cite{JMLR:v21:20-074}, as a text-to-text transfer transformer, simplified the training procedure and improved the performance of previous models. 
\begin{figure*}[!t]
    \centering
    \includegraphics[width=.9\linewidth]{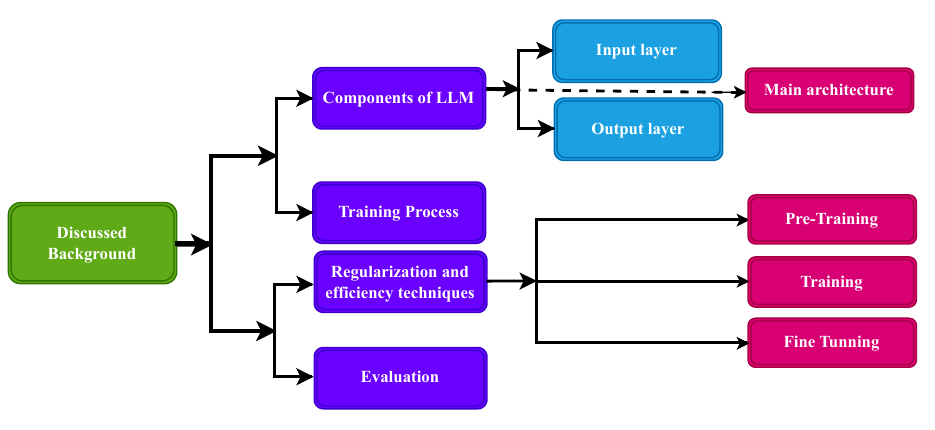}
    \caption{Update it based on context, make it smaller maybe, correct font.}
    \label{LLL_model}
\end{figure*}
Later on in 2020, GPT-3~\cite{NEURIPS2020_1457c0d6} with 175 billion parameters showed an impressive ability in question answering and text generation. DALL-E~\cite{pmlr-v139-ramesh21a} and~\ac{CLIP}~\cite{Radford2021LearningTV}, introduced in 2021, are the first multi-modal models that are able to generate images from texts and match images with texts. 
\ac{ChatGPT} was introduced in 2022 by adding conversational and user interaction capabilities to~\ac{GPT}-3.~\ac{GPT}-4, with its ability to understanding complex and long texts, handling ambiguity, advanced reasoning and multi-modal capability is the latest~\ac{GPT} model in 2023.
By analyzing different large language models, we find out that the inseparable and key components of every large language model are the main structure, pre-training, training and fine-tuning techniques, evaluation metrics, as well as the regularization methods  
Each one of the state-of-the-art methods tries to improve the performance of~\ac{LLM} by addressing several limitations in these main components, as we tend to discuss them in more details in this section ( See Fig. \ref{LLL_model}).
\subsection{Components of~\ac{LLM}}
Large language models and specifically~\ac{ChatGPT} models, as complex deep neural networks, compose of several key components in their structures, which enables them to understand and generate human languages. From general point of view, the structure of~\ac{LLM} consists of data pre-processing part~\cite{raiaan2024review}, that prepares text data for the inferring or training stages, and input layers~\cite{naveed2023comprehensive}, which are responsible to convert raw data
into data vectors that are understandable by the model, the main architecture of the model, and the output layer, which prepares the final results to be understandable by human~\cite{hadi2023large}.
\subsubsection{Data Pre-processing}
Data pre-processing step is a critical phase which ensures the consistency and compatibility of the raw text data to be used in the~\ac{LLM}. Text cleaning, handling long sequences of data, and text augmentation are the main pre-processing techniques in~\ac{LLM}.
\textbf{Text cleaning} is mainly responsible to remove noises such as irrelevant and non-necessary information, ensure uniformity of the input data, reduce the text complexity and enhance data integration. These goals can be obtained by removing special characters that are irrelevant to the semantic meaning of the the text~\cite{Jurafsky2009}, unifying the text by lower casing the letters and removing additional white spaces~\cite{raiaan2024review}, and handling contractions by expanding them into their full formats~\cite{manning2008introduction}.
\textbf{Handling long sequence} of data should also be done before using raw data within the~\ac{LLM} due to limitations of computational resources. This goal can be reached by truncating long sequences that exceeds a specific length~\cite{Devlin2019BERTPO}, splitting the long chains of words into smaller ones~\cite{Liu2019RoBERTaAR}, or using sliding windows to use small overlapped sequence of words~\cite{Dai2019TransformerXLAL}. 
\textbf{Text augmentation techniques}~\cite{feng-etal-2021-survey} such as synonym replacement~\cite{wei-zou-2019-eda}, contextual augmentation~\cite{kobayashi-2018-contextual}, back translation~\cite{sennrich-etal-2016-improving} enhance the diversity of the data which itself results in training a more generalized model and improves the performance of the model. 
\subsubsection{Input layers}
After the pre-processing phase, the raw text data is converted into a suitable format by input layers in such a way that it can be processed by~\ac{LLM}. 
\textbf{Tokenization}~\cite{Webster1992TokenizationAT}, as the first input layer, is responsible to split sequences of words into smaller units called tokens, based on which the model will be trained to understand the human language. Different levels of tokenization can be utilized in~\ac{LLM}~\cite{Mielke2021BetweenWA}. 
Character-level~\cite{mcnamee2004character, ribeiro2018study} tokenization is the simple form of tokenization which consider each character in the text as a separate token. Although this method can handle any characters in any type of text, it cannot represent semantic information properly and it will provide large sequences of data for the training phase~\cite{wang2024tokenization}. Word-level tokenization~\cite{Mikolov2013EfficientEO,Mielke2021BetweenWA} tries to split text into words based on the space and punctuation between words. Although this method is an instinctive tokenization technique, it encounter the ~\ac{OOV}~\cite{moon2021effects} problem where the model faces a new word in the testing phase that does not exist in the training phase. To deal with~\ac{OOV} problem, sub-work tokenizer~\cite{kudo2018subword,Kudo2018SentencePieceAS} is introduced which splits words into smaller units to handle the rare and unseen words issue. ~\ac{BPE}~\cite{sennrich-etal-2016-neural}, WordPiece~\cite{zhang2016google}, and SentencePiece~\cite{kudo-richardson-2018-sentencepiece} are known as the most common sub-word tokenizers in~\ac{LLM}. It is good to mention that~\ac{GPT} series utilize different modification of BPE as their tokenizer methods.
\textbf{Embedding layer} is another input layer which is mainly responsible for converting input tokens into vector representations that can be processed by the model and can capture linguistic properties during the training phase~\cite{naveed2023comprehensive}. Generally, one can divide the embedding into two groups as token embedding~\cite{raiaan2024review}, where each unique token from the tokenizer will be mapped to a high-dimensional vector to capture the semantic meaning of the token, and positional embedding or encoding~\cite{vaswani2017attention} which tends to provide the model with the position of each token inside a sequence of data. Token embedding techniques ~\cite{patil2023survey} are related to the tokenizer types as they try to convert characters, words, and sub-words into related vectors. On the other hand, positional encoding can be divided into several techniques~\cite{lopez2024positional} such as~\ac{APE} which assigns a unique vector to each position like~\ac{BERT}~\cite{devlin2018bert}, ~\ac{LPE} in which the positional embeddings are learned in the training phase like~\ac{GPT} series~\cite{NEURIPS2020_1457c0d6},~\ac{RPE} that encodes the distance between the tokens like T5~\cite{JMLR:v21:20-074} and Transformer-XL~\cite{Dai2019TransformerXLAL}, and~\ac{RoPE} which encodes the positions in the self attention mechanism like in RoFormer~\cite{su2024roformer}.
With these input layers, the raw data would be ready to be processed by the main architecture of the~\acp{LLM} Which we tend to introduce in more details.
\subsubsection{Main architecture}
Although~\ac{LLM} are built on different architecture principles~\cite{Minaee2024LargeLM}, the core of most modern ones, like~\ac{GPT} series, are Transformers~\cite{vaswani2017attention}. From one point of view, transformers in~\ac{LLM} may be only an encoder, a decoder, or an encoder-decoder frameworks~\cite{Minaee2024LargeLM}. 
The encoder-based models like BERT~\cite{devlin2018bert} can be used for feature extraction and context understanding as they capture the important patterns of the input data and learn the dependencies between the input sequences properly. Encoder-based models are suitable to be used for text classification and sentiment analysis~\cite{devlin2018bert}, text summarization~\cite{liu-lapata-2019-text} and text similarity detection~\cite{wang-etal-2018-glue}. On the other hands, decoder-based~\ac{LLM} like~\ac{GPT} series are mainly used for text generation. Decoder-encoder~\ac{LLM} are the most common~\ac{LLM} which are designed to convert input sequences into output sequences and are suitable to be used in dialogue generation, translation, and text summarizing applications~\cite{hadi2023large}. 
It is also good to mention that encoders and decoders in~\ac{LLM} consist of several main layers to reach their goals. Different types of attention mechanism~\cite{9194070}, ~\ac{FFN}, layer normalization, and residual connections are the most important ones~\cite{naveed2023comprehensive}.
\textbf{Attention mechanisms}~\cite{de2022attention} play a crucial role in~\ac{LLM} as they try to capture the dependencies between input tokens while emphasizing on more relevant tokens during the training process. Different attention mechanisms are developed to improve the contextual understanding of the model and enhance the performance of the model in complex and multi tasks applications~\cite{Chaudhari2019AnAS}. Self attention or intra-attention~\cite{vaswani2017attention} is one of most common attention techniques that can be used in both encoder or decoder frameworks to figure out the relationship between different positions of input tokens or target sequences. Cross attention or global attention~\cite{gheini-etal-2021-cross} is mainly used in encoder-decoder architectures to find the dependencies between different positions of encoder and decoder sequences. Local attention~\cite{luong-etal-2015-effective} mainly focuses on finding the dependencies of subset of sequences in the decoder part by avoiding the expensive computations. Sparse attention~\cite{Child2019GeneratingLS} utilizes sliding windows to reduce the computations of self attention method in the encoder and decoder frameworks. Flash attention~\cite{Dao2022FlashAttentionFA} is a memory efficient attention technique compatible to be used with~\acp{GPU}.
\textbf{Feed-forward neural networks}, as simple but important layers, contribute significantly to the overall performance of the~\acp{LLM} by adding non linearity to encoders or decoders and transferring features specifically in the multilingual language models~\cite{bhattacharya2024understanding}.
\textbf{Normalization layers}~\cite{vaswani2017attention,ba2016layer}, as inseparable part of~\ac{LLM} architectures, are responsible for stabilizing, accelerating the training process, and improving the generalization~\cite{pmlr-v119-xiong20b}. Moreover, residual connections or skip connections~\cite{vaswani2017attention} facilitate the training process by addressing the vanishing gradient problems and improving the convergence properties~\cite{kobayashi-etal-2021-incorporating}.
One can say that the above-mentioned layers are used in encoder or decoder architecture of almost every modern~\ac{LLM} as each one of them improves the performance of the model in a specific way.
\subsubsection{Output layer}
The projection and softmax layers are the last layers of LLMs that are responsible to map the output vector to the token space, produce logits for each token, and provide a probability distribution over the vocabulary for the next token~\cite{vaswani2017attention}.
\subsection{Pre-Training, Training and Fine Tuning}
Large language models are developed during a multi-phase process as pre-training, training and fine-tuning which the performance of the models depends on directly~\cite{liu2024understanding}. Different techniques are used in each one of these steps to fulfill specific objectives which we tend to discuss in more details in this section. 
\subsubsection{Pre-training and Training}
Pre-training is the first and important step of developing~\acp{LLM} which allows the model to learn the general structures of the language from a diverse corpus of raw text data gathered from various datasets and places such as Internet~\cite{liu2024understanding}. Pre-training in~\acp{LLM} is an unsupervised process as no labeled data or specific guidance is used in this manner~\cite{Minaee2024LargeLM}. In this step, a general-purpose model is trained in such a way that it can capture semantic and syntactic information of the text and generate contextually reasonable texts across different topics and styles based on previous data. 
This phase creates a generalized model which performs well in unseen data and prepares a robust model that can be improved further in the training and fine-tuning phases for specific applications. 
Many methods such as~\ac{MLM}~\cite{Devlin2019BERTPO},~\ac{CLS}~\cite{Radford2019LanguageMA}, Mixture of Experts~\cite{shazeer2017outrageously,fedus2022switch}, and Sequence-to-Sequence (seq2seq) Modeling~\cite{lewis2019bart} have been developed to pre-train~\acp{LLM} efficiently~\cite{Minaee2024LargeLM}. 
\textbf{~\ac{MLM}} techniques like ~\ac{BERT}~\cite{Devlin2019BERTPO}, mask some parts of the input sequences while the model tries to predict the masked part based on the surrounding context considering the future and past tokens. These pre-training methods are mainly used in applications like text classification, sentiment analysis, and named entity recognition. 
\textbf{~\ac{CLS}} is an auto-regressive pre-training technique~\cite{radford2018improving} in which the model tries to predict the next word based the previous tokens. It is well-suited for applications involving summarization and text generation and is widely used in the pre-training of~\ac{GPT},~\ac{GPT}-2, and~\ac{GPT}-3 models. 
\textbf{~\ac{MoE}} pre-training techniques, like Switch Transformers~\cite{fedus2022switch} and GShard~\cite{lepikhin2020gshard}, improve the efficiency and scalability of the model as they divide the model into several neural networks known as experts, and selectively activate a subset of these experts for each input token, allowing better utilization of the model capacity and resources without a proportional increase in computational cost~\cite{shazeer2017outrageously}.
\textbf{Sequence to sequence} techniques, like BART~\cite{lewis2019bart}, T5~\cite{JMLR:v21:20-074}, and MarianMT~\cite{junczys-dowmunt-etal-2018-marian}, work based on an encoder-decoder architecture where the decoder generates the output sequence from the input token processed by the encoder. These methods are particularly useful for complex applications that involve input-output transformations such as machine translation, summarization, and conversational AI. There are also other pre-training techniques like Denoising Autoencoders~\cite{yang2019xlnet}, in which tokens are permuted randomly and the model is trained to predict the original order resulting in robust pre-training, or D4~\cite{tirumala2024d4}, which utilizes document de-duplication and diversification techniques to speed up the pre-training process and improves the downstream accuracy. The most important thing in all of these methods is that they try to reduce the computational costs of the pre-training phase while creating a robust and generalized model for further steps.
It is also good to mention that few methods utilize a training step, also known as continued pre-training, in addition to the pre-training phase in order to enhance the model's understanding by utilizing additional or more specific datasets~\cite{liu2024understanding}. Although this step creates a more robust model with updated knowledge on a specific language by integrating the new datasets that were not included in the pre-training phase, it requires more computational resources, as a result of which many methods skip this phase and try to use more efficient approaches in the fine-tuning step.
\subsubsection{Fine-tuning}
\label{sec:fine-tuning}
As it is mentioned above, fine-tuning~\acp{LLM} to improve their performance for specific tasks is a crucial phase in developing them. Although few literature reviews~\cite{Minaee2024LargeLM} have been tried to categorize some of these methods from specific points of views, there is not comprehensive introduction of these methods. Due to this reason, we tend to give a more comprehensive categorization of these methods.
We divide fine-tuning methods into five different groups as supervised techniques, ~\ac{PEFT}, instruction approaches, alignment techniques, and safety fine-tuning methods and try to introduce them in more details in this section.
\textbf{Supervised fine-tuning} methods utilize a labeled dataset to optimize the pre-trained language model to perform better on a specific task. Many approaches like transfer learning, multi-task learning, task specific learning, and few-shot learning are some the most common supervised approaches. In the task-specific fine-tuning methods like~\cite{zheng2024fine} and BERT~\cite{Devlin2019BERTPO}, the model parameters are optimized based on a specific dataset, developed for a target task, in such a way that the model excels in generating specific content with precision and accuracy. Transfer learning methods like~\cite{rehan2023fine} optimize a previously trained model on a task to be transferred to another related task. Multi-task fine-tuning technique like~\cite{chen2021multi}, sharing representation simultaneously across different tasks to improve the accuracy. In the few-shot learning approaches like~\cite{brown2020language,schick2020exploiting}, the pre-trained model is optimized by learning effectively from just a few examples of the new task which is useful when only few labeled data for a specific task is available.
\textbf{Instruction tuning}~\cite{zhang2023instruction} is another type of~\acp{LLM} fine-tuning techniques which involves fine-tuning the model for various tasks while following specific commands provided by human. Multi-turn instruction following, self-instruction, and natural instructions tuning are the most common techniques in this category. Natural instruction methods~\cite{mishra-etal-2022-cross} utilizes a dataset including a variety of instructions expressed in natural language instructions, allowing the model to learn from these instructions and use them in inference phase. The InstructGPT~\cite{ouyang2022training}, as the best example of natural instruction methods, fine-tunes~\ac{GPT}-3 to follow human instructions. Self-instruction technique~\cite{wang-etal-2023-self-instruct} tries to generate structures from a language model, filter similar or invalid ones, and then use them for fine-tuning the model. On the other hands, multi-turn instruction techniques~\cite{hu2024fine,10.1145/3651671.3651702} handle instructions across multiple conversational turns, while keeping context, and providing comprehensive responses throughout the interactions. These methods can be used in applications like conversational agents, customer support systems, and interactive dialogue systems.
\textbf{~\ac{PEFT}} methods~\cite{xu2023parameter} have become popular in~\acp{LLM} due to huge number of trainable parameters in these models and the limitation of computational resources for the training purposes. Generally,~\ac{PEFT} methods can be categorized into five groups as additive, unified, re-parameterized, hybrid, and partial fine-tuning techniques. Additive methods involve strategies where additional parameters are defined and fine-tuned while the majority of the original model's parameters are kept frozen.  Some of the most recently developed additive approaches are CoDA~\cite{lei2023conditional}, MerA~\cite{he2023mera}, and AdapterSoup~\cite{chronopoulou2023adaptersoup} while Sequential Adapter~\cite{houlsby2019parameter} and Residual Adapter~\cite{lin-etal-2020-exploring} are the firstly developed techniques. All of these methods reduce the computational and memory requirements significantly, making it possible to improve the performance of~\acp{LLM} in various applications.
Partial fine-tuning methods like US-BitFit~\cite{ding2023parameter} and SAM~\cite{fu2023effectiveness} involve only a subset of the model's parameters affecting the downstream tasks while keeping the rest fixed. Re-parameterized are the most common fine-tuning methods that reduce the number of trainable parameters by utilizing low rank transformations or additional structures. LoRA~\cite{hu2021lora} and its extensions~\cite{xu2023parameter} are the most efficient re-parameterized techniques that improve generalization and enhance performance. Hybrid fine-tuning methods like AutoPEFT~\cite{zhou2024autopeft} combine different~\ac{PEFT} techniques to utilize advantages of several methods while discarding their limitations. Last but not least, unified approaches like ProPETL~\cite{zeng-etal-2023-one} involve various fine-tuning methods into a single architecture ensuring consistency across different methods. Unified methods has only a single~\ac{PEFT} architecture unlike hybrid ones that utilize several~\ac{PEFT} architectures during the fine-tuning process. 
It is good to mention that~\cite{xu2023parameter} introduce state-of-the-art~\ac{PEFT} methods in more details discussing about their pros and cons.
\textbf{Alignment tuning} methods are responsible to lay out a broad framework for lining up~\acp{LLM} with human values, ethical guidelines, and specific user requirements, ensuring their safe and effective deployment~\cite{li2024agent}. Since~\acp{LLM} are pre-trained on massive corpus data from different datasets, they may generate biased, harmful, unreliable, and unethical contents. The main purpose of alignment methods is to make sure that the generated content by~\acp{LLM} is harmless, honest, and helpful~\cite{naveed2023comprehensive}. The most common alignment method is Reinforcement Learning from Human Feedback (RLHF)~\cite{ouyang2022training} in which human feedback is used to fine-tune the model by defining a Reward Modeling (PM) or Comparative Ranking (CR), and an optimization algorithm such as Proximal Policy Optimization (PPO)~\cite{schulman2017proximal} to optimize the rewards. Reinforcement Learning from AI Feedback (RLAIF)~\cite{lee2023rlaif} is a more recent alignment method which aims to improve the alignment of models by using feedback generated by AI systems rather than human feedback alone. Other alignment methods such Direct Preference Optimization~\ac{DPO}~\cite{rafailov2024direct} tries to find a different mapping approach between the reward functions and optimal policies solving the instability and complexity challenges of RLHF techniques. 
\textbf{Safety Fine-tuning} ~\ac{SFT}~\cite{raza2024developing} is a critical part of~\acp{LLM} to prevent harmful outputs in real-world applications where safety and ethical considerations are paramount. Supervised Safety Fine-Tuning which uses high safety risk adversarial data in the~\ac{SFT} process, Safety RLHF~\cite{NEURIPS2020_1f89885d} utilizing a safety reward model within the~\ac{RLHF} framework, and Safety Context Distillation~\cite{askell2021general} converting safety preprompts to adversarial prompts in fine-tuning process, are some of the~\ac{SFT} approaches in~\acp{LLM}~\cite{liu2024understanding}. 
\subsection{Regularization and Efficiency Techniques}
\subsubsection{Efficiency Methods}
These techniques in~\acp{LLM} focus in reducing the computational costs, memory requirements, and latency in the training, fine-tuning and inferring phases without compromising the accuracy and performance significantly. We categorized these methods into several groups as model quantization, model pruning, knowledge distillation, structure optimization, parallelism, parameter sharing, memory scheduling, and low-rank approximation.
\textbf{Model Quantization} techniques~\cite{jacob2018quantization} reduce the number of floating-point bits in model weights and decrease the precision of the numbers in numerical calculations which itself reduces the memory usage and speeds up the training and inferring phases in~\acp{LLM}. ~\ac{PTQ}~\cite{jacob2018quantization}, which is applied to an~\ac{LLM} after the training phase, and~\ac{QAT}~\cite{esser2019learned}, integrating the quantization into the training phase during both forward and backward processes, are the most common quantization techniques. 
There are also other quantization approaches like Per-Tensor and Per-Channel Quantization~\cite{krishnamoorthi2018quantizing} which limits the quantization to a specific group parameters, Mixed-Precision Quantization~\cite{micikevicius2017mixed} that uses different precision levels for different parts of the model, Binary and Ternary Quantization~\cite{courbariaux2016binarized} which reduces weights to very low bit-widths. 
Although these methods reduces the memory and energy consumption, and speed up the inference process, they can compromise the accuracy due to squeezing of high precision weight to lower precision ones. To deal with these issues, many recent approaches use a combination of these techniques like BinaryBERT~\cite{bai-etal-2021-binarybert} that trains the model with a ternary model and then a binary model through splitting its weights.
\textbf{Model Pruning} methods~\cite{sun2023simple} reduce the model size and computational requirements by removing less important and redundant parameters in~\acp{LLM}. Generally, pruning techniques are divided into structured and unstructured methods. Unstructured methods aim to remove connections or parameters without considering any specific structural patterns. Wanda~\cite{sun2023simple}, which removes unimportant weights in every level based on the norm of input, ~\ac{OWL}~\cite{yin2024outlierweighedlayerwisesparsity}, which is an extension of Wanda by adding layer pruning to it, and ~\ac{CAP}\cite{xu2022dense}, which prunes the model based on contrastive loss between the pre-trained and fine-tuned model, are some of the recently developed unstructured pruning methods. On the other hand, structured methods prune models based on a structural patterns~\cite{Gordon2020CompressingBS}.~\ac{LLM}-Pruner~\cite{ma2023llm}, which prunes non important coupled structures based on gradient information, Bonsai pruner~\cite{dery2024everybody}, that is a gradient-free pruning model with only forward passes, and an optimization-based structural pruner~\cite{gao2024optimization} that works based on Bernoulli distribution, are some of the recent structured pruning methods. 
\textbf{Knowledge Distillation} methods~\cite{hinton2015distilling} are student-teacher based models in which the knowledge is transferred from a complex model (teacher) to a smaller model (student) which maintains performance while being more efficient.~\cite{xu2024survey} represents a comprehensive survey on different knowledge distillation methods by categorizing them into three different groups as KD algorithms, skill distillation, and verticalization distillation methods. KD algorithms are responsible for training a student model to reproduce the behavior of a teacher model while skill distillation techniques aim to transfer specific capabilities from teacher to student models. On the other hand, verticalization distillation tries to transfer knowledge across different levels of abstraction within the LLMs.
\textbf{Structural Optimization} efficiency techniques aim to optimize the structure of the components of~\acp{LLM} to reduce the memory access operations and increase the performance respectively. As an example of these methods, FlashAttention~\cite{dao2022flashattention} and PagedAttention~\cite{kwon2023efficient} try to improve computational speed by using a chunked computation method, which reduces the memory requirements in~\ac{SRAM} typically associated with matrices. On the other hand, NoMAD-Attention~\cite{zhang2024nomad} utilizes Single-~\ac{SIMD} registers in CPU to enhance the efficiency of the~\acp{LLM} during the inference phase by solving the challenges of~\ac{MAD} matrix operations in the attention computations.
\textbf{Parallelism} techniques enhance training and inferring efficiency by distributing the model across multiple processors and enabling different model components to be processed concurrently. Parallelization approaches mainly appear in four categories~\cite{brakel2024model,liu2024understanding} as data parallelism, model parallelism, pipeline, and mixed parallelism. Data parallelism~\cite{10.5555/3433701.3433727,ren2021zero,zhao2023pytorch} involves in dividing the data into smaller batches and splitting these batches among several processors. Finally, the gradients of data in the related batches are summed up and used to update the model. Although data parallelism is easily implemented and scales nicely with the number of parameters, updating model parameters and aggregating gradients requires effective synchronization techniques.
On the other hand, in model parallelism, the model itself is split across multiple processors~\cite{shoeybi2019megatron}. It is good to mention that, tensor parallelism refers to a type of model parallelism in which the parameters of the model are divided into many tensors, each of which is computed on a separate processing unit~\cite{brakel2024model}. Although model parallelism techniques are very effective for certain tasks, such as matrix multiplications, they are difficult to use and need sophisticated inter-processor communication in order to transfer intermediate results.
In pipeline parallelism~\cite{huang2019gpipe,ao2024seq1f1b}, device utilization is improved by vertically extending the number of GPU units through parallel computing to support larger models. Although pipeline parallelism distributes memory and computation workload among processors, it needs careful synchronization to handle the data transfer between phases. Mixed parallelism~\cite{narayanan2021efficient} combines data, model, and pipeline parallelism to take use of each approach's advantages. Although mixed parallelism optimizes hardware resource consumption, it requires complicated implementation and tuning to achieve ideal performance.
\textbf{Parameter Sharing} techniques~\cite{lan2019albert,su2024beyond}, such as tying the weights of different layers or using recurrent structures, aim to reduce the number of trainable parameters by sharing a common set of weights across different parts of the model. Weight sharing improves the computational efficiency of the model and lowers the chance of over fitting, particularly when there is little data available. 
\textbf{Memory Scheduling} methods~\cite{kwon2023efficient,sun2024llumnix,han2022bminf} refers to the effective arrangement and control of memory access patterns throughout the decision-making or inference stage. Large models frequently have intricate structures and significant memory requirements when utilized in sophisticated reasoning tasks like natural language processing or complex decision-making. By optimizing the retrieval and storage of intermediate representations, model parameters, and activation values, memory scheduling ensures that the inference process is accurate and runs as quickly as possible.
\subsubsection{Regularization Methods}
These techniques also increase the efficiency of the~\acp{LLM} by preventing over fitting, improving convergence, and enhancing the model's ability to generalize to unseen data. Some of the most common and effective regularization techniques in~\acp{LLM} are dropout layers~\cite{srivastava2014dropout}, layer-wise dropout~\cite{ni2024layer}, early stopping techniques~\cite{dodge2020fine}, and gradient clipping~\cite{lee2021scaling,li2021large}. Other methods such as Mixout regularization~\cite{lee2019mixout} and privacy regularization~\cite{mireshghallah-etal-2021-privacy} are recently developed to obtain specific goals in large language models.
\subsection{Evaluation}
Evaluation of language models specifically large ones is one the most challenging and complex tasks as it requires a thorough analysis of the model considering different issues and aspects~\cite{liu2024understanding}. The evaluation procedure measures the model's capability to understand, produce, and communicate in a variety of contexts using human language. It involves assessing the model's performance, identifying limitations, addressing ethical concerns, and ensuring technical robustness~\cite{raiaan2024review}. Continuous evaluation and improvement are essential to harness the full potential of~\acp{LLM} while mitigating their risks and drawbacks.
As the evaluation process is an inseparable part of~\acp{LLM} development, we discuss the importance of evaluation and introduce different evaluation metrics in details in this section.
\subsubsection{Evaluation Tasks}
Language model evaluation tasks can be categorized into three main groups as~\ac{NLU}, ~\ac{NLG}, and security evaluations.~\ac{NLU} evaluation tasks~\cite{karanikolas2023large} tend to assess understanding performance of the language model. They evaluate a variety of tasks, including as text classification, sentiment analysis, natural language inference, question answering, mathematical reasoning, commonsense reasoning, and reading comprehension. On the other hand,~\ac{NLG} evaluation tasks~\cite{gao2024llm} evaluate ~\acp{LLM}' text generating capabilities by comprehending the input context that has been supplied. It covers activities including conversation creation, sentence completion,~\ac{MT}, and summarization. 
Security evaluation becomes an inseparable part of any~\acp{LLM} nowadays as~\acp{LLM} have to be aware of any potential security risks, stop malicious use or attack weaknesses, and deal with any long-term problems that can endanger human development. Generally speaking, security evaluation should analyze the levels of potential bias, privacy protection, and adversarial attacks in the LLMs~\cite{liu2024understanding}.~\acp{LLM} face some potential biases in the training data and analyzing how accurate they can deal with these biases is an essential evaluation task. The causes and consequences of potential biases in~\acp{LLM}, such as~\ac{ChatGPT}, are discussed in~\cite{ferrara2023should} and the possible approaches to reduce them is discussed in~\cite{gehman-etal-2020-realtoxicityprompts}. On the vother hand, privacy protection refers to the protection of user data that is used during the training and inferring phases to avoid data misuse. A comprehensive research~\cite{nasr2023scalable} on~\ac{ChatGPT} shows that the user data, such as text and image, used in the training phase can be extracted and misused. Although the methods like~\ac{DEPN}~\cite{wu-etal-2023-depn} can be applied to reduce the privacy leakages, it is important to evaluate the levels of data privacy in~\acp{LLM} precisely.
Adversarial attacks~\cite{zou2023universal} such input tampering, deliberate disinformation, or fake information generation could also affect the performance of~\acp{LLM}. When evaluating security, it is important to take into account how resilient the model is to this kinds of assaults~\cite{zhang2022remos}. 
\subsubsection{Evaluation Metrics}
Evaluation metrics should be developed in such a way that they can assess the LLMs from different points of views as is discussed in the previous section. Accuracy, hallucination, robustness, reasoning, fairness, and generalizations are some of the most important criteria that evaluation metrics should be able to analyze~\cite{hu2024unveiling}. Accuracy and reasoning~\cite{jin2023cladder} refers to fact that how accurate the model can grasp the most important information from the inputs and generate the appropriate and correct outputs. Hallucination~\cite{ganguli2022red,ji2023survey} refers to the fact that if information in an~\ac{LLM} output is fabricated or false while robustness~\cite{goyal2023survey} explains how an~\ac{LLM} can withstand mistakes while still producing reliable and consistent results. Generalization~\cite{hupkes2023taxonomy} and fairness~\cite{hovy2021importance} demonstrates an~\ac{LLM}'s capacity to adjust to unseen data, which is essential for answering a variety of questions and comprehending text production processes.
Although evaluation metrics are categorized based on their applications into three groups as~\ac{MC},~\ac{TS}, and~\ac{QA} metrics in~\cite{hu2024unveiling}, we categorize them based on their structure into five groups as character-based, word-based, embedding-based, language model-based, and~\ac{LLM} assisted metrics. 
\textbf{Character-based} metrics focus on the individual characters and their sequences, which can be important for tasks involving non-word entities, spelling, or languages with complex character systems. Edit-distance~\cite{przybocki2006edit} as the most common character-based metrics determines the minimum number of single-character adjustments (insertions, deletions, or replacements) needed to transform a word or text string into another that can be helpful when assessing spelling corrections or tasks in which fine-grained text accuracy is crucial.
\textbf{Word-based} metrics evaluate the quality of the generated text at the word level, providing insights into the model's linguistic capabilities.~\ac{BLEU},~\ac{ROUGE}, and ~\ac{METEOR} are some the most common word-based metrics.~\ac{BLEU}\cite{papineni2002bleu} compares the~\ac{LLM}'s output to annotated ground truths by measuring the precision of n-grams in the generated text against one or more reference texts. ROUGE~\cite{lin2004rouge}, which is a common method to evaluate text summaries from~\ac{NLP} models, calculates recall by measuring the overlap of n-grams, word sequences, and word pairs between the generated text and reference texts. METEOR~\cite{banerjee2005meteor} is a more comprehensive metric than the previous ones as it utilizes a combination of n-gram matches and n-grams overlaps, adjusted for word order differences between the outputs of~\acp{LLM} and the reference data. 
It is also important to keep in mind that character and word-based metrics are not accurate enough to assess the long and complex outputs of~\acp{LLM} as they have limited reasoning ability and cannot take into account semantic information of the data properly.
\textbf{Embedding-based} metrics rely on word or sentence embedding to measure the quality of the generated text, focusing on its semantic meaning rather than individual characters. BERTScore and MoverScore are the most common embedding-based metrics. BERTScore~\cite{zhang2019bertscore} evaluates the performance of the model by computing the cosine similarity between the contextual embedding of words in the ground truth data and the generated texts extracted from the pre-trained BERT model. On the other hand, MoverScore~\cite{zhao2019moverscore} evaluates the model, by computing the minimum cost required to transform the word embedding of the generated text into the word embedding of the reference text extracted by a pre-trained model like~\ac{BERT}. Although BERTScore and MoverScore represent a significant advancement in the evaluation of text generation models, their dependencies on contextual embeddings from trained models makes them vulnerable to contextual awareness and biases.
\textbf{Language Model-based} metrics rely on NLP models to solve the limitations of the embedding-based metrics.~\ac{NLI},~\ac{BLEURT}, and~\ac{QAQG} are some of the most common model-based metrics. Entailment or~\ac{NLI} metric~\cite{naveed2023comprehensive} utilizes the inference abilities of a natural language model to analyze how much the generated~\ac{LLM} output is logically entailment, contradictory, or neutral compared to a ground truth text.~\ac{BLEURT}~\cite{sellam-etal-2020-bleurt} presents a methods for combining expressivity and robustness that involves pre-training a fully learned metric on a substantial quantity of synthetic data and then fine-tuning it using ratings from humans. On the other hand,~\ac{QAQG}~\cite{honovich-etal-2021-q2} metric is used to assess the consistency of the generated text with the reference data. The process begins with creating pairs of questions and answers from a reference text and then comparing the answers, generated by the model for the same set of questions, to the reference answers. Although these metrics give a better evaluation for the~\acp{LLM}' performance, they sometimes struggle with accuracy while dealing with long and complex texts.
\textbf{~\ac{LLM}-assisted} metrics utilize other LLMs' outputs to evaluate the performance of a specific large model. GPTScore, SelfCheckGPT, G-Eval, and Prometheus are some the most LLM-based metrics. GPTScore~\cite{fu2023gptscore} as the first~\ac{LLM}-assisted metrics aims to evaluate multiple aspects of generated text such as informativeness and relevancy. This method relies on the assumption that higher-quality generations will have higher token probabilities assigned by the~\ac{LLM} while trying to utilize the conditional likelihood of generating the target text as an assessment metric. Similar to GPTscore, G-EVAL~\cite{liu-etal-2023-g} also evaluates the generated text by using large language models such as~\ac{GPT}-4 with the chain-of-thoughts framework; however, in contrast to GPTscore, assessment is carried out directly by advising the model to assign a score to the generated text. SelfCheckGPT~\cite{manakul2023selfcheckgpt} is a fact-check evaluation metric that aims to detect and quantify hallucinations without a reference text. This method utilizes the basic idea that sampling responses are likely to be comparable and contain consistent information if an~\ac{LLM} is familiar with a particular subject and makes use of this idea to detect hallucination. Prometheus~\cite{kim2023prometheus} and its extensions Prometheus2~\cite{kim2024prometheus} are completely open-source~\ac{LLM} that utilize~\ac{GPT}-4 to create a new dataset and use that to evaluate any long-form text based on a user-customized score. It is clear that~\ac{LLM}-assisted metrics can give a better evaluation of the model due to their complexity and the way that they can analyze the output from different perspective mentioned before.
Now that you are familiar with the main components of~\acp{LLM}, their training and fine-tuning methods, regularization and efficiency techniques as well as evaluation metrics, we tend to discuss about the architecture and functionality of~\ac{GPT} models in details in the next section. 
\begin{comment}
{\color{red}Historical Context: Review the development of~\acp{LLM} leading up to ChatGPT.\\
Technical Foundations: Explain the underlying technologies, such as neural networks and transformers.\\
Key Terminology: Define essential terms and concepts used throughout the paper.}
\end{comment}
\section{\ac{ChatGPT}: Architecture and Functionality }\label{Archi}
\begin{figure*}[!t]
    \centering
    \includegraphics[width=1\linewidth]{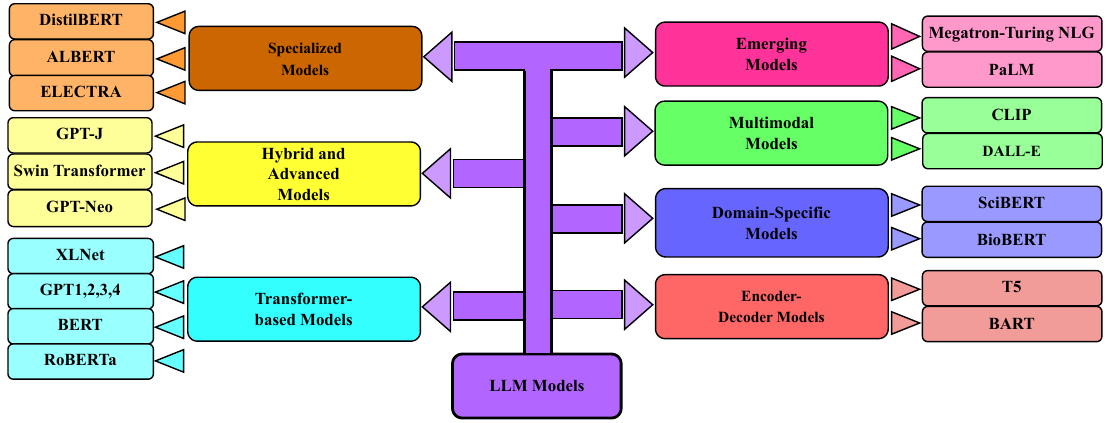}
    \caption{Different LLM Models}
    \label{Different_LLM_Models}
\end{figure*}

\ac{ChatGPT}, a prominent example of a large language model~\ac{LLM}, is built on the transformer architecture, which forms the foundation for its ability to understand and generate human-like text. At its core,~\ac{ChatGPT} utilizes the~\ac{GPT} series developed by OpenAI, specifically leveraging advancements from models like GPT-3 and~\ac{GPT}-4. These models are pre-trained on diverse datasets encompassing vast amounts of text from the internet, enabling them to grasp intricate language patterns and context. The transformer architecture's self-attention mechanism allows~\ac{ChatGPT} to assess the importance of different words in a sentence, facilitating nuanced and coherent responses. Various iterations of~\ac{GPT} have progressively enhanced the model's capabilities, with each version incorporating larger datasets and more parameters, thereby refining its performance in tasks such as translation, summarization, and question-answering. The flexibility and scalability of the transformer architecture make it possible for~\ac{ChatGPT} to adapt and improve continuously, addressing complex language tasks with remarkable accuracy and fluency. This section delves into the architectural components and functionalities of~\ac{ChatGPT}, highlighting how different~\ac{LLM} models contribute to its sophisticated language generation capabilities (See Fig. \ref{Different_LLM_Models}).
\begin{comment}
{\color{red} foscusing on:\\
Comparison with Other~\acp{LLM}\\
Feature Comparison: Compare ChatGPT with other prominent~\ac{LLM} (e.g., BERT, T5).\\
Performance Metrics: Analyze performance in various benchmarks and tasks.\\
Innovations and Improvements: Highlight unique features and advancements of ChatGPT.\\}
\end{comment}
\subsection{Transformer-based Models} 
 \subsubsection{~\ac{GPT} Series}
The~\ac{GPT} series by OpenAI includes advanced language models like~\ac{GPT}-2,~\ac{GPT}-3, and ~\ac{GPT}-4, designed for understanding and generating human-like text. Pre-trained on vast datasets, these models excel in tasks such as email spam detection, translation, medical, information extraction, summarization, and question-answering, significantly advancing natural language processing capabilities \cite{mohammad2023large}. \\
To counteract numerous advanced malware attacks, threat actors are now leveraging~\ac{GPT} and other~\acp{LLM} to devise sophisticated strategies for system infection with new malware. The authors in \cite{shandilya2023gpt} aim to showcase specific methods that can be employed to mitigate the risks associated with malware created using~\ac{ChatGPT} and other~\ac{LLM}-based tools. \\
The study \cite{luo2023self}, written in an accessible format, introduces the transformer architecture. The authors explain how this innovative design efficiently processes long sequences and captures relationships over great distances. \\
The authors in \cite{yan2023real} present an~\ac{OPF} model that integrates linguistic stipulations, proposing a novel solution method by incorporating a~\ac{GPT}-based~\ac{LLM} agent into the primal-dual~\ac{DRL} training loop. This approach allows for the direct modeling of traditionally unquantifiable linguistic stipulations, expressed in natural language, as objectives and constraints in the OPF problem. The~\ac{GPT}-agent converts the satisfaction of these linguistic stipulations into corresponding rewards and constraints. These non-differentiable rewards, produced by the~\ac{GPT}-agent, are refined and optimized through continuous interactions with the environment during the~\ac{DRL} process. Once the~\ac{DRL} agent is adequately trained, it can solve the~\ac{OPF} model in real time. \\
Parl et. al, in \cite{park2024lpddr} present CXL-PNM, a novel processing near-memory platform leveraging~\ac{CXL} technology to accelerate Transformer-based~\acp{LLM}. The authors highlight the unique trade-offs between bandwidth and capacity offered by LPDDR5X memory modules, which are well-suited for handling the increasing complexity of LLMs. Their setup delivers 2.9 times greater energy efficiency, 31$\%$ higher throughput and 23$\%$ lower latency. Compared to a~\ac{GPU}-based system with 8~\ac{GPU} devices, their designed model reduces hardware costs by 30$\%$.~\ac{GPT} series are as follows:  
\begin{itemize}
\item~\ac{GPT}-1: The original model introduced by OpenAI, demonstrating the potential of unsupervised pre-training followed by supervised fine-tuning.
\item~\ac{GPT}-2: A more powerful version with 1.5 billion parameters, known for its ability to generate coherent and contextually relevant text.
\item~\ac{GPT}-3: Significantly larger with 175 billion parameters, capable of performing a wide range of tasks with minimal fine-tuning.
\item~\ac{GPT}-4: The latest iteration, offering even greater capabilities and improvements in performance and fine-tuning efficiency.
\end{itemize}
The authors in \cite{jeong2024advancing} introduce the innovative use of~\acp{LLM} in tinnitus therapeutics to analyze~\ac{CBT} and predict treatment outcomes, thereby aiding in the management of high patient caseloads. By anonymizing patient data and applying~\ac{GPT}-2-based embeddings, along with dimensionality reduction and clustering techniques, they observe how patients' misconceptions change and their emotional discomfort decreases. Their clustering results demonstrate that~\acp{LLM} can provide valuable insights into~\ac{CBT} processes. To address the limitations posed by a small dataset, they augment the textual patient data in three ways, incorporating a penalty to minimize augmentation bias. This augmented data is used to train the Google T5 Transformer, enabling it to predict~\ac{THI} score outcomes at the conclusion of~\ac{CBT} sessions. \\
The authors in \cite{maddigan2023chat2vis} introduce an innovative method called Chat2VIS, which harnesses the capabilities of pre-trained~\acp{LLM} such as~\ac{ChatGPT} and~\ac{GPT}-3 to transform free-form natural language into executable code for generating visualizations. Chat2VIS demonstrates that, through the use of strategically designed prompts,~\acp{LLM} can reliably create visualizations from natural language queries, even when these queries are vague or poorly specified. This method not only significantly reduces the costs associated with developing~\ac{NLI} systems but also achieves superior visualization inference capabilities compared to traditional~\ac{NLP} methods that rely on hand-crafted grammar rules and specialized models. Furthermore, their research outlines the construction of~\ac{LLM} prompts in a manner that ensures data security and privacy, while maintaining generalizability across various datasets. \\
The authors in \cite{mai2023data} delve into an innovative data augmentation methodology that leverages a pretrained~\ac{LLM}, specifically OpenAI~\ac{GPT}-3.5 Turbo, for generating new data and filtering high-quality data for final use. Their study centers on a~\ac{NLI} task in the Vietnamese language, encompassing four labels: "contradiction", "neutral", "entailment"  and "other". Diverging from conventional methods that typically involve word substitution or deletion, their approach harnesses the~\ac{LLM}'s capabilities to completely rewrite sentences according to prompts tailored for each label definition. \\
Contemporary datasets often suffer from significant obsolescence or deficiencies in both quality and quantity due to organizations' reluctance to share data, driven by concerns over privacy or the potential compromise of proprietary information. To tackle this issue, the authors in \cite{kholgh2023pac} present PAC-GPT, an innovative framework designed to generate reliable synthetic data for machine learning applications, leveraging OpenAI's Generative Pre-trained Transformer 3 (GPT-3). Central to this framework are two key modules: a Flow Generator, tasked with capturing and reproducing patterns within sequences of network packets, and a Packet Generator, capable of generating individual network packets based on the network flow. Additionally, they propose a packet generator employing~\ac{LLM} chaining and subsequently assess, compare, and evaluate its performance using metrics such as loss, accuracy, and success rate. Their findings indicate that transformers offer a viable approach for synthetic packet generation, requiring minimal fine-tuning. \\
In \cite{danner2023advancing}, a novel~\ac{AI} application is introduced for detecting depression, utilizing advanced transformer networks to analyze clinical interviews. By integrating simulated data to supplement traditional datasets, the authors address issues surrounding data protection and privacy, thereby enhancing the model's effectiveness. Their method utilizes BERT-based models,~\ac{GPT}-3.5, and~\ac{ChatGPT}-4, resulting in cutting-edge results in depression identification through linguistic patterns and contextual clues, surpassing previous techniques.\\
The authors in \cite{nascimento2023gpt} introduce the '~\ac{GPT}-in-the-loop' methodology, aiming to explore the cognitive abilities of LLMs such as ~\ac{GPT} within~\ac{MAS}. Departing from traditional adaptive strategies that often entail lengthy training periods, their framework leverages~\ac{GPT}-4 to bolster problem-solving and explanatory proficiencies. To investigate this methodology, they implement it in a smart streetlight scenario within the~\ac{IoT} framework, where each streetlight is managed by an autonomous agent fitted with sensors and actuators, assigned with the task of establishing an energy-efficient lighting scheme. By incorporating~\ac{GPT}-4, these agents exhibit improved decision-making and flexibility, eliminating the need for extensive training. \\
The authors in \cite{eloundou2023gpts} explore the potential ramifications of~\acp{LLM}, exemplified by~\acp{GPT}, on the U.S. job market, honing in on the augmented functionalities resulting from~\ac{LLM}-driven software vis-à-vis standalone~\acp{LLM}. Employing a fresh assessment framework, they gauge professions based on their alignment with~\ac{LLM} capabilities, amalgamating both human expertise and~\ac{GPT}-4 categorizations. The anticipated consequences traverse all income tiers, with affluent occupations potentially encountering more extensive exposure to~\ac{LLM} functionalities and~\ac{LLM}-powered software. Importantly, these repercussions extend beyond sectors with heightened recent productivity growth.\\
The authors in \cite{hu2024teaching} study the pedagogical content knowledge theory, which initially formulates an instructional design framework based on mathematical problem sequences and corresponding prompt instructions. Subsequently, they develop a comprehensive tool for assessing~\ac{LLM}'s instructional design capabilities . A dataset for high school mathematics teaching plans is generated using Generative Pretrained Transformer 4. Eventually, the efficiency of~\acp{LLM} in instructional design is examined that reveals the teaching plans generated by~\acp{LLM} excel in various aspects. Some of those aspects are selecting methods and strategies, organizing problem chains, identifying teaching priorities, teaching activities, setting instructional objectives, and articulating subject content. \\
The study \cite{ramprasad2024analyzing} is the first to thoroughly evaluate the performance of LLMs in dialogue summarization, and it reveals significant discrepancies that highlight the continued difficulties in this field. The study's conclusions highlight the frequency of circumstantial inferences in the summaries produced by~\ac{GPT}-4 and Alpaca-13b, demonstrating LLMs' ability to comprehend English but propensity to add conceptual interpretations. Furthermore, the authors show that current measurements are unable to accurately identify these subtle inaccuracies. 
 \subsubsection{~\ac{BERT}} 
\begin{itemize}
\item~\ac{BERT} Base: Trained on masked language modeling and next sentence prediction tasks, it captures context from both directions in the text.
\item~\ac{BERT} Large: A larger version of BERT with more parameters, providing improved performance on various~\ac{NLP} tasks.
\end{itemize}
~\ac{BERT}'s intrinsic token limit of 512 tokens creates a considerable obstacle when processing very long documents, a frequent scenario in legal document reviews where documents often surpass this limit. To address this issue, the authors in \cite{yang2023empirical} empirically evaluate two different strategies for applying~\ac{BERT}, using real-world data from the construction industry. Based on the first strategy,~\ac{BERT} is applied to the entire document in its entirety. On the other hand, the second strategy divides each document into smaller text portions and then applies~\ac{BERT} to these small texts separately. After a comparative evaluation, the most effective approach for handling long legal documents is selected. \\ 
As these models advance,~\ac{PaLM},~\ac{BERT}, and~\ac{GPT} are anticipated to become increasingly proficient in topic categorization. The authors in \cite{yerramreddy2023empirical} conduct a comparative analysis of~\ac{PaLM},~\ac{BERT}, and~\ac{GPT} language models for topic categorization, utilizing the AG News dataset. Their results demonstrate that the proposed models offer more precise categorization for the text within the dataset. \\
~\ac{FL} is a distributed~\ac{ML} approach that stores the data locally on the devices, allowing several devices to collaborate to train a model while preserving privacy. Instead of broadcasting raw data, the devices broadcast model updates (such gradients) to a central server, which is the fundamental idea behind ~\ac{FL}. Together, these updates are then added to the global model~\cite{AKHTARSHENAS2024107964}.
The study \cite{sadot2023novel} proposes a novel approach for building a ML model for multilabel classification of large textual datasets using~\ac{FL}. This method involves extending~\ac{BERT}-based structure with a one-dimensional~\ac{CNN}. Initially, the experiment is conducted on a single machine with the entire dataset. Next, the dataset was divided into two subsets, and the experiment was repeated in a~\ac{FL} setup.~\ac{FL} setup is significantly  power-efficient while improving F1 score, precision, and accuracy. \\
Training~\acp{LLM} in the medical field is challenging due to restricted data access and privacy regulations such as the~\ac{GDPR} and~\ac{HIPAA}.~\ac{FL} offers a collaborative learning while safeguarding data security. The authors in \cite{peng2024depth} evaluate~\ac{FL} on two biomedical natural language processing tasks across eight datasets using six different~\ac{LLM}. They show that~\ac{FL} models consistently surpass models trained on individual client data and achieve significantly better results than pre-trained~\ac{LLM} using few-shot responding techniques.
The study \cite{miyashita2024bert} presents an innovative approach for conducting movie keyword searches by leveraging user-generated rankings and reviews. The authors harness the capabilities of the~\ac{BERT} language model, fine-tuned specifically for this task. Their model is trained to grasp the intricate relationships between keywords and movies using paired data from user-generated movie rankings and reviews collected from a prominent Japanese movie review platform. With a dataset encompassing 10,000 user rankings and 15,000 films, the model shows higher efficiency that traditional similarity-based methods in a binary classification task, resulting in superior performance. \\
To enhance the product’s quality, cost, customer service, and environmental impact, managers aim to analyze customer ratings and the underlying emotional content of reviews . For example, the proposed research \cite{ishtiaq2024product} uses the~\ac{LLM} to accurately predict product helpfulness . This assists customers in saving time and money. To develop numerous advanced ML tools, they employ a benchmark dataset, the Amazon Fine Food Reviews. Introducing a novel transformer approach called~\ac{BERF} (BERT Random Forest) for feature engineering aimed at improving the value of user evaluations for Amazon’s gourmet food products. They use the Synthetic Minority Over-sampling Technique (SMOTE) approach to balance the dataset. \\
It's crucial to determine the veracity of news provided on Twitter. In any language, including Turkish, it is crucial to identify fraudulent tweets. To create a dataset called TR-FaRe-News, \cite{koru2024detection} uses the Zemberek natural language processing tool, which is designed for the Turkish language, to preprocess and label fake news sourced from Twitter verification platforms. Next, word2vec, TF-IDF, and collective algorithms are used to investigate the R-FaRe-News dataset in order to detect fake news. \\
The writers of \cite{fu2024iov} discuss how conventional security approaches are unable to handle the intricate and varied connections seen in the Internet of Vehicles.~\ac{IoV}-~\ac{BERT}-~\ac{IDS}, the suggested framework, uses~\ac{BERT} to obtain an~\ac{IoV}-wide representation of raw traffic data. By using less data for fine-tuning and unsupervised pre-training, this method improves accuracy. The semantic extractor converts unprocessed data into contextual semantic traffic pairs, so addressing the problem of unclear semantics in traffic data. Moreover, the inclusion of two pre-training tasks adds bidirectional contextual features to~\ac{IoV}-~\ac{BERT}-~\ac{IDS}, which greatly aids in the recognition and learning of contextual features and models. \\
 \subsubsection{XLNet}
 An extension of Transformer-XL that incorporates permutation-based training, capturing bidirectional context without the limitations of masked language modeling.\\
XLNet is an extended~\ac{AR} pretraining technique that combines the benefits of~\ac{AE} and~\ac{AR} techniques with a permutation language modeling aim. By carefully developing the two-stream attention mechanism and integrating Transformer-XL, the neural architecture of XLNet is engineered to function in unison with the~\ac{AR} goal. On a variety of activities, XLNet significantly outperforms earlier pretraining targets \cite{yang2019xlnet}. \\ 
 \subsubsection{~\ac{RoBERTa}}
 An optimized version of BERT, trained with more data and longer sequences, leading to improved performance on benchmark tasks.\\
In \cite{liu2019roberta}, the authors assess several design choices during the process of pretraining~\ac{BERT} models. They discover that training the model longer, with larger batches over more data, training on longer sequences, eliminating the next sentence prediction target, and continuously modifying the masking pattern applied to the training data may lead to improved performance. Their study results in the~\ac{GLUE},~\ac{SQuAD}, and~\ac{RACE} with their improved pretraining technique, which they called~\ac{RoBERTa}, without the need for extra data for SQuAD or multi-task finetuning for GLUE. \\
Researchers are "fine-tuning"~\ac{BERT} and~\ac{RoBERTa} models to perform better on tasks that are specific to their domains. For example, the authors in \cite{hong2022scholarbert} assess how changes along different dimensions (e.g., training data, model size, pretraining time, finetuning length) impact downstream performance by applying 14 transformer-based models to 11 scientific tasks. By training a 770M-parameter BERT model on a 221B token scientific literature dataset spanning multiple disciplines, they produce ScholarBERT, the largest and most diverse scientific language model to date. \\
\subsection{Encoder-Decoder Models}
\subsubsection{~\ac{T5}}
It treats every~\ac{NLP} task as a text-to-text problem, enabling a unified approach to diverse tasks such as translation, summarization, and question answering. It also uses an encoder-decoder architecture to convert text-to-text for various tasks, providing flexibility and consistency across different~\ac{NLP} tasks.\\
The authors of \cite{raffel2020exploring} present a uniform framework that translates all text-based language problems into a text-to-text format, allowing readers to explore the landscape of transfer learning techniques for~\ac{NLP}. Pre-training objectives, architectures, unlabeled data sets, transfer methodologies, and other aspects are compared on numerous of language understanding problems in their systematic study. Through the integration of their newly acquired "Colossal Clean Crawled Corpus" and scale, along with the insights gleaned by their research, they are able to attain cutting-edge outcomes on numerous metrics, including text categorization, question answering, summarization, and more. \\
The authors of \cite{jeong2024advancing} demonstrate how an~\ac{LLM}-based strategy can precisely forecast the results of the tinnitus~\ac{CBT} treatment and greatly lessen the effort involved in evaluating each~\ac{CBT} session. The findings of their investigation show that even with a dataset full of grammatical and arithmetic errors, Google~\ac{T5} and its variant Flan-~\ac{T5} ~\ac{LLM} can predict the right treatment results. This emphasizes how solid and dependable the suggested strategy is. Finally, given their encouraging findings in their investigation, they advise adopting highly trained~\ac{LLM} models, such as Google~\ac{T5} and Flan-~\ac{T5}~\acp{LLM} with supplemented datasets, for this clinical adaptation. \\
%%acro here
In \cite{gong2024ast}, the authors provide~\ac{AST}-~\ac{T5}, a brand-new pretraining paradigm that improves code creation, translation, and comprehension by utilizing the~\ac{AST}. Their~\ac{AST}-Aware Segmentation preserves code structure through adaptive programming, whereas our~\ac{AST}-Aware Span Corruption goal gives the paradigm the tools it needs to reassemble other code structures. Since~\ac{AST}-~\ac{T5} does not require intricate program analyses or architectural modifications, it may be integrated with any encoder-decoder Transformer, in contrast to other models. Test results reveal that~\ac{AST}-~\ac{T5} often beats LMs of comparable size in a variety of code-related tasks, such as MBPP and HumanEval. In code-to-code tasks,~\ac{AST}-~\ac{T5} outperforms CodeT5 due to its structure-awareness. \\
The new development of deep learning and~\ac{LLM} can greatly aid in the critical task of summarizing medical reports so that the general public can easily access them. To summarize these reports, the authors in \cite{helwan2023medical} suggest an improved~\ac{T5}. Indiana Dataset, which is accessible to the public, is used to train and test their proposed model and is eventually assessed with the ROUGE collection of measures. \\
\subsubsection{~\ac{BART}}
It combines bidirectional and autoregressive training, effective for text generation and transformation tasks like summarization and translation. \\
Introduced in \cite{lewis2019bart}, BARTs are pre-training techniques that learn to map corrupted texts to the original.  In several text creation tasks,~\ac{BART} delivers new state-of-the-art outcomes while performing comparably to~\ac{RoBERTa} on discriminative tasks. \\
Due to their good performance,~\acp{PTM} are now widely available. Model selection usually takes the model's paradigm, like auto-recursive, Encoder Decoder, Masked Language Modeling into account. Therefore, a more appropriate~\ac{BART} model can be chosen as a strong tool for the summary assignment \cite{liu2024understanding}. \\
\subsection{Specialized Models}
\subsubsection{DistilBERT}
A smaller, faster, and more efficient version of~\ac{BERT}, trained using knowledge distillation to retain much of~\ac{BERT}’s performance with fewer parameters. \\
The authors of \cite{sanh2019distilbert} provide a technique for pre-training DistilBERT, a smaller general-purpose language representation model that can be fine-tuned to perform well on a variety of tasks similar to its larger counterparts. Although the majority of earlier research focused on using distillation to create task-specific models, they also used distillation of knowledge in the pre-training stage. In order to take advantage of the inductive biases that larger models pick up during pre-training, their model incorporates a triple loss that combines cosine-distance, distillation, and language modeling.
\subsubsection{~\ac{ALBERT} or A Lite~\ac{BERT}}
An efficient version of BERT with parameter-sharing and factorized embedding parameterization to reduce memory consumption and increase training speed. \\
~\ac{ALBERT}-xxlarge produces far better results than~\ac{BERT}-large with fewer parameters, however, due to its larger structure, it requires more computing power. Therefore, increasing~\ac{ALBERT}'s training and inference speed using techniques like block attention and sparse attention is a crucial next step. Another research area with greater representation power is the study of hard example mining and more effective~\ac{LLM} training \cite{lan2019albert}. Furthermore, the authors foresee the possibility of additional dimensions not yet captured by the current self-supervised training losses. These dimensions could enhance the representational power of the resulting models. Despite this, they have compelling evidence that sentence order prediction is a consistently useful learning task. This task produces better language representations. \\
\subsubsection{~\ac{ELECTRA}}
Uses a discriminator-generator setup for pre-training, where the discriminator learns to distinguish between real and replaced tokens, providing efficient and effective training. \\
In \cite{ni2022electra}, the~\ac{ELECTRA} model for prompt-based zero-shot learning on~\ac{NLP} tasks is examined. A brand-new~\ac{RTD}-based prompt learning technique is suggested by the authors. Through rigorous trials on 15 different datasets, the~\ac{ELECTRA} model performs unexpectedly well as a few shot and zero shot learner, their model indicates that it has greater room for stimulation. Their~~\ac{RTD}-\ac{ELECTRA}-large, for example, performs an astounding 90.1\% zero-shot on the SST-2 challenge. It may learn more pre-knowledge because of the well-designed~\ac{RTD} pre-training challenge, which is mostly responsible for its higher efficiency. Furthermore, their study demonstrates~\ac{ELECTRA}'s excellence as a zero-shot learner. \\
\subsection{Multimodal Models}
\subsubsection{\ac{CLIP}}
Trained on image and text pairs, capable of understanding and generating descriptions for images, as well as performing zero-shot classification. \\
It has been demonstrated that contrastive models such as~\ac{CLIP} are able to learn stable representations of images that capture both style and meaning. The authors in \cite{ramesh2022hierarchical} suggest a dual approach to take advantage of these representations for image generation: a prior that, given a text caption, provides a~\ac{CLIP} image embedding, and a decoder that, in turn, generates an image conditioned on the image embedding. They demonstrate how improving image variety may be achieved with no compromise in photorealism or narrative similarity when image representations are actively generated. By altering the non-essential information missing from the image representation, their decoders conditioned on picture representations can also generate versions of an image that maintain its semantics and style. \\
The efficacy of various vision encoders inside~\acp{MLLM} is thoroughly investigated by the authors in \cite{jiang2023clip}. Their results show that~\ac{CLIP}'s shallow layer features are especially beneficial for fine-grained tasks like grounding and area understanding. Remarkably, as a visual branch inside~\acp{MLLM}, the vision-only model~\ac{DINO}, which is not pretrained with text-image alignment, performs excellently. In fine-grained linked perception tasks,~\ac{DINO} outperforms~\ac{CLIP} just by adding an MLP layer for alignment. They also suggest COMM, a powerful feature merging technique that combines Multi-level features merging with~\ac{CLIP} and~\ac{DINO} to improve~\ac{MLLM}s' visual capabilities. \\
The~\ac{CLIP} training paradigm restricts the exposure of different texts to the same image by applying data augmentations only to the image inputs and leaving the language inputs unaltered throughout the training process. The authors of \cite{fan2024improving} present~\ac{LaCLIP}, a straightforward but incredibly powerful method for improving~\ac{CLIP} training via language rewrites. They rebuild the text descriptions linked to each image by utilizing huge language models' in-context learning capacity. These revised texts maintain the main ideas and meanings while showcasing a variety of language and sentence structures.~\ac{LaCLIP} chooses the modified or original sentences at random to use as text augmentations for every image during training. \\
\subsubsection{DALL-E}
DALL-E is a useful text-to-image model that generates images from textual descriptions, showcasing the ability of~\acp{LLM} to create coherent and contextually relevant visual content~\cite{ramesh2021zero} . Therefore, to interact with AI to carry out activities, people employ foundation models like text-to-image models DALL-E and~\acp{LLM}~\ac{GPT}-4. Despite the fact that users can access foundation models via chatbots (like~\ac{ChatGPT}), chat is not a production tool for creating repeatable~\ac{AI} services, regardless of the power of the underlying models. While~\acp{API} such as LangChain make~\ac{LLM}-based application development possible, they also provide a barrier because they demand a high level of programming competence. \\
In order to address the aforementioned issue, the authors of \cite{cheng2023prompt} systematize the methodology for~\ac{AI} chain engineering by methodically reviewing, summarizing, improving, and expanding the idea of~\ac{AI} chain by incorporating the best practices and principles that have been amassed over decades in software engineering. Additionally, they create Prompt Sapper, a no-code integrated development environment that organically incorporates these ~\ac{AI} chain engineering principles and concepts into the building process, enhancing the effectiveness and caliber of~\ac{AI} chains. \\
DALL-E's composition-based systematic generalization ability in picture generation has demonstrated remarkable results; nevertheless, it needs a dataset of text-image pairs, and the text provides the compositionality. On the other hand, models of representation that are object-centric, such as the Slot Attention model, acquire composable representations without requiring guidance from text. But, in contrast to DALL-E, it has far less systematic generalization capacity for zero-shot generation. In order to address this issue, the authors of \cite{singh2021illiterate} suggest
%afshin {\color{red}SLATE1},
a slot-based autoencoding architecture that combines the best features of both approaches: object-centric representation learning that enables methodical generalization in text-free zero-shot image synthesis. Another way to think about their paradigm is as an illiterate DALL-E model. They employ the Image GPT decoder conditioned on the slots for capturing intricate interactions between the slots and pixels, in contrast to the pixel-mixture decoders of the current object-centric representation models. \\
The~\ac{SLD} framework, a ground-breaking self-correction system that uses detectors and~\acp{LLM} to greatly improve text-to-image alignment, was described by the authors in \cite{wu2024self}. This technique is compatible with a number of generative models, including DALL-E 3, and it also establishes a new~\ac{SOTA} in the image generating benchmark. Additionally,~\ac{SLD} expands its usefulness to picture editing programs by providing more precise object-level manipulation than current techniques. \\
\subsection{Domain-Specific Models}
\subsubsection{~\ac{BioBERT}}
Pre-trained on biomedical text, optimized for tasks like named entity recognition, relation extraction, and question answering in the biomedical domain.  \\
A domain-specific language representation model pre-trained on extensive biomedical corpora is called~\ac{BioBERT}, according to the authors in \cite{lee2020biobert}. Pre-trained on biomedical corpora,~\ac{BioBERT} performs significantly better than~\ac{BERT} and previous state-of-the-art models across a range of biomedical text mining tasks, all while maintaining almost the same structure within workloads.~\ac{BioBERT} significantly surpasses~\ac{BERT} on three representative biomedical text mining tasks: biomedical named entity recognition (0.62\% F1 score improvement), biomedical relation extraction (2.80\% F1 score improvement), and biomedical question answering (12.24\%~\ac{MRR} improvement). \\
\subsubsection{~\ac{SCIBERT}}
Trained on scientific literature, it is designed to handle tasks specific to scientific text processing and understanding. \\
The authors in \cite{beltagy2019scibert} provide~\ac{SCIBERT}, a pretrained~\ac{BERT}-based language model for scientific text. We assess SCIBERT using a variety of datasets and tasks from scientific fields. On multiple of these tasks,~\ac{SCIBERT} obtains new SOTA outcomes and surpasses~\ac{BERT}-Base greatly, even matching some published~\ac{BioBERT} results on biomedical problems. \\
\subsection{Hybrid and Advanced Architectures}
\subsubsection{~\ac{GPT}-Neo}
It is an open-source alternative to~\ac{GPT}-3, and offers powerful language generation capabilities with varying parameter sizes.\\
\ac{LLM}-based algorithms for code evolution have just lately become popular in the field of~\ac{GP}. A codified~\ac{LLM}-based evolutionary algorithm called~\ac{LLM}-~\ac{GP} is presented by the authors in \cite{hemberg2024evolving} with the purpose of evolving code. It employs evolutionary operators, just like~\ac{GP}, but its designs and implementations of those operators are very different from~\ac{GP}'s since they take advantage of an~\ac{LLM}'s pre-trained pattern matching and sequence completion skills as well as prompting. \\
In \cite{zhang2024map}, the authors present MAP-Neo, a fully open-source multilingual~\ac{LLM} suite that takes significant steps to improve the transparency and accessibility of large language models~\acp{LLM}. They aim to assist the academic and open-source communities in furthering transparent~\ac{NLP} research by publishing in-depth details of their procedures, which range from pre-training corpus (e.g., Matrix Data Pile), data curation, model training, and evaluation. \\
\subsubsection{Swin Transformer}
A vision transformer model designed for image recognition tasks, capable of handling high-resolution images efficiently.
Transformer's widespread attention has led to an increase in its attention in the field of computer vision. For example, the authors \cite{xu2024swin} combine the benefits of~\ac{ResNet} with the~\ac{Swin} Transformer to build the Swin Transformer and~\ac{ResNet}-based ~\ac{STRN} for low-light image improvement. Multiscale discriminators and a U-shaped generator make up the~\ac{STRN}. Three modules make up the generator: one for deep feature extraction, one for shallow feature extraction, and one for picture reconstruction. In deep learning approaches, users can also employ ResNet and~\ac{Swin} Transformer blocks to calculate global and local attention. The random paired training of~\ac{STRN} is constrained by the self perceptual loss and the spatial consistency loss. \\
~\ac{Swin} Transformer, a novel vision Transformer with linear computing complexity with respect to input image size, is presented by the authors in \cite{liu2021swin}. It generates a hierarchical feature representation. Swin Transformer far outperforms earlier top techniques to attain state-of-the-art performance on~\ac{COCO} object detection and ADE20K as a semantic segmentation dataset. The authors anticipate that Swin Transformer's impressive results on a range of vision-related issues will promote a unified analysis of language and visual signals. \\
A new model, called~\ac{SwinIR}, is proposed by the authors in \cite{liang2021swinir}. The shallow feature extraction, deep feature extraction, and high resolution reconstruction modules make up the three main components of the framework. Specifically, for deep feature extraction, they employ a stack of~\ac{RSTB}, which are made up of a residual connection, a convolution layer, and~\ac{Swin} Transformer layers. According to their research,~\ac{SwinIR} performs at the cutting edge in three typical image restoration jobs. \\
\subsection{Emerging Models}
\subsubsection{Megatron-Turing NLG}
This model was developed by NVIDIA as a massive transformer model designed for natural language generation, aiming to push the boundaries of language model performance.\\
Large-scale training of~\acp{LLM} with billions of parameters is difficult and takes a lot of processing power. In order to extract this computation from Frontier, the first exascale supercomputer in the world devoted to open science, the study \cite{dash2024optimizing} investigates effective distributed training methodologies. To hide or reduce latency, the authors must choose the best mix of distributed and parallelization strategies to overlap communication and computation. To this purpose, they migrate cutting-edge distributed training frameworks like Megatron-DeepSpeed and FSDP to Frontier, where they set up a software stack for training~\ac{LLM} models. 
To make training a trillion-parameter model on Frontier easier, they enable and explore multiple model and data parallel training strategies, including sharded data parallelism, pipeline parallelism, and tensor parallelism. \\
The authors in \cite{smith2022using} provide details on the training of Megatron-Turing NLG 530B~\ac{MT-NLG} with 530 billion parameters, as the product of a collaborative effort between Microsoft and NVIDIA. Using DeepSpeed and Megatron, they trained this model in 3D parallelism, with an initial emphasis on the infrastructure. The training procedure, their training corpus's architecture, and their data curation methods—which they consider to be essential to the model's performance—are then covered in detail.  \\
\subsubsection{\ac{PaLM}}
Developed by Google, a large-scale model leverages the Pathways system to improve efficiency and performance across diverse tasks.\\
In today's information-rich digital environment, topic classification is essential in many ways. While several models are presented to classify the themes using different datasets, \ac{PaLM},\ac{BERT}, and\ac{GPT} models are examples of the \ac{LLM} that have been developed recently. With increased accuracy, these models can detect the subjects mentioned in a document because they can understand the statistical correlations between words and sentences. Furthermore, they have a fast text data processing speed, which is crucial for applications that need real-time subject categorization \cite{yerramreddy2023empirical}. \\
The authors in \cite{chowdhery2023palm} developed a 540-billion parameter\ac{PaLM} to deepen their understanding of the effect of scale on few-shot learning. They use Pathways, a novel machine learning framework that allows for highly efficient training across several\ac{TPU} Pods, to train\ac{PaLM} on 6144\ac{TPU} v4 chips. By attaining state-of-the-art few-shot learning outcomes on hundreds of language understanding and generation benchmarks, they continue to highlight the benefits of scalability.\ac{PaLM} 540B demonstrates breakthrough performance on several of these tasks, including a series of multi-step reasoning problems where it outperforms the refined state-of-the-art.~\ac{PaLM} excels at generating source code and multilingual activities as well. \\
\section{Various dataset of~\ac{LLM}}\label{dataset}

\begin{table*}[ht]
\caption{List of datasets for the Pre-Training stage of~\acp{LLM}.}
\centering
\begin{tabular}{p{2.3cm}p{2.5cm}p{12cm}}
\hline
 \textbf{Dataset Domain} & \textbf{Dataset Type} & \textbf{Dataset Names and applied paper}     \\
\hline
 General Purpose & Web Data & Common Crawl~\cite{crawl2019common}, CC-Stories~\cite{trinh2018simple}, CLUECorpus~\cite{xu2020cluecorpus2020}, C4~\cite{raffel2020exploring}, mC4~\cite{Xue2020mT5AM}, RefinedWeb~\cite{10.5555/3666122.3669586}, WuDaoCorpora~\cite{yuan2021wudaocorpora}, WanJuan-CC~\cite{Qiu2024WanJuanCCAS}, cc100~\cite{conneau-etal-2020-unsupervised}, RedPajama-V2~\cite{together2023redpajama} \\
\hline
 General Purpose & Books and Literature & BookCorpusOpen~\cite{bookcorpusOpen}, PG-19~\cite{rae2019compressive}, Toronto Book~\cite{zhu2015aligning}\\
\hline
General Purpose & Academic Materials & PubMed, S2ORC~\cite{lo-etal-2020-s2orc}  \\
\hline
 General Purpose & Social Media Content  & OpenWebText~\cite{Gokaslan2019OpenWeb}, Pushshift Reddit~\cite{baumgartner2020pushshift}, WebText~\cite{radford2019language} \\
\hline
General Purpose & Code  & BIGQUERY~\cite{Nijkamp2022CodeGenAO}, phi-1~\cite{gunasekar2023textbooks}, The Stack~\cite{kocetkov2022stack} \\
\hline
General Purpose & Encyclopedia Content &WikiMatrix~\cite{schwenk-etal-2021-wikimatrix}, Wikipedia~\cite{wikipedia}, TigerBot-wiki~\cite{Chen2023TigerBotAO}  \\
\hline
 General Purpose & Multi-category & ROOTS~\cite{laurenccon2022bigscience}, The Pile~\cite{pile}, Dolma~\cite{soldaini2024dolma}, MAP-CC~\cite{du2024chinese}, RedPajama-V1~\cite{together2023redpajama}, TigerBot-pretrain~\cite{Chen2023TigerBotAO}, WanJuanText~\cite{he2023wanjuan}, SlimPajama~\cite{cerebras2023slimpajama}, Minerva~\cite{lewkowycz2022solving} \\
\hline
General Purpose & Conversational Data & DailyDialog~\cite{li2017dailydialog}, ConvAI~\cite{logacheva2020convai2} \\
\hline
 General Purpose & Legal Documents  &EuroParl Corpus~\cite{koehn2005europarl}  \\
\hline
 Domain Specific & Multilingual Content  &CulturaX~\cite{nguyen-etal-2024-culturax}, OSCAR~\cite{abadji-etal-2022-towards}, MADLAD-400~\cite{10.5555/3666122.3669062}, TigerBot~\cite{Chen2023TigerBotAO}  \\
\hline
 Domain Specific & Multi-modal Data &mOSACR~\cite{futeral2024moscar}  \\
\hline
 Application Specific & Finance &BBT-FinCorpus~\cite{lu2023bbt}, FinCorpus~\cite{zhang2023xuanyuan}, TigerBot-earning~\cite{Chen2023TigerBotAO}, TigerBot-research~\cite{Chen2023TigerBotAO}  \\
\hline
 Application Specific & Medical &Medical-pt, PubMed Central  \\
\hline
 Application Specific & Math & OpenWebMath~\cite{paster2023openwebmath}, MathPile~\cite{wang2023generative}, Proof-Pile-2~\cite{azerbayev2023llemma} \\
\hline
 Application Specific & Law & TigerBot-law~\cite{Chen2023TigerBotAO} \\
\hline
 Application Specific & Transportation & TransGPT-pt~\cite{TransGPT} \\
\hline
\end{tabular}
\label{dataset:pretrain}
\end{table*}

In the development of~\acp{LLM}, datasets are typically categorized based on their role in the model’s life cycle as pre-training, fine-tuning, and evaluation~\cite{liu2024datasets}. Different datasets are used in each of these stages to reach specific goals and obtain optimal model performance. Although many recent studies~\cite{liu2024datasets, naveed2023comprehensive, Minaee2024LargeLM} have tried to categorize these datasets from different points of view, they do not have comprehensive recent lists of dataset used in each stage, as well as the fact that they do not represent datasets based on their applications in specific details. In this section, we proposed a novel perspective to categorize~\acp{LLM}' datasets, mainly considering the applications, tasks, and domains of~\acp{LLM} as well as the types of the methods used for developing them.  

\subsection{Datasets for Pre-Training}
Datasets, used for the pre-training phase, can be categorized into three groups: general purpose, domain, and application specific datasets~(See Tab.~\ref{dataset:pretrain}). The general purpose datasets are collected from different resources such as web pages, books, news, academic materials, codes, social media, etc., based on which they are categorized in Tab.~\ref{dataset:pretrain}. Their key feature, which makes them suitable for training generic foundational models, is that the text content is collected from various domains. The large amount of data and its diversity allows~\acp{LLM} with huge amount of parameters to be trained properly in the pre-training stage.
Common Crawl~\cite{crawl2019common} and its recent extensions such as CC-Stories~\cite{trinh2018simple}, CLUECorpus2020~\cite{xu2020cluecorpus2020}, C4~\cite{raffel2020exploring}, mC4~\cite{Xue2020mT5AM}, RefinedWeb~\cite{10.5555/3666122.3669586} are some of the most common general dataset where large corpus of data is collected from different web pages. Moreover, recent general purpose datasets such as ROOTS~\cite{laurenccon2022bigscience}, Pile~\cite{pile}, and Dolma~\cite{soldaini2024dolma} try to collect data from different resource to train a more generalized model during the pre-training phase. We categorized them as the multi-category data resources in Tab.~\ref{dataset:pretrain}.
On the other hand, datasets gathered for particular applications are known as application-specific ones which are utilized gradually in the pre-training phase of~\acp{LLM} for specific applications such as mathematics, medicine, law, finance, and transportation.
Domain-specific datasets are the ones that belong to a specific domain and share common features while being used in different applications. Multilingual datasets such as TigerBot~\cite{Chen2023TigerBotAO} and CulturaX~\cite{nguyen-etal-2024-culturax} which contain text in various languages, and Multi-modal datasets like mOSACR~\cite{futeral2024moscar} containing different types of data, are the most recently developed domain-specific datasets.

It is also good to mention that some~\acp{LLM}, like~\ac{GPT} models, utilize different resources during theirpre-training,g which enables them to produce better contextualized information across various domains. The main data resource in all~\ac{GPT} series is large web page data, while~\ac{GPT}-1 and~\ac{GPT}-2 also utilize data from books and news;~\ac{GPT}-3 and~\ac{ChatGPT} use literature, news, scientific, and conversation data in addition to the public web data~\cite{raiaan2024review}. 
\begin{table*}[ht]
\caption{List of datasets for the Fine-Tuning stage of~\acp{LLM}.}
\centering
\begin{tabular}{p{2.7cm}p{3.3cm}p{9.8cm}}
\hline
 \textbf{Dataset Domain} & \textbf{Dataset Type} & \textbf{Dataset Names and applied paper}     \\
\hline
%1
 Instruction Fine-Tuning & Human Generated &OASST1~\cite{wang2023openchat}, Aya~\cite{singh2024aya}, InstructIE ~\cite{gui2023instructie} \\
\hline
%2
 Instruction Fine-Tuning & Model Constructed &CAMEL~\cite{li2023camel}, LMSYS-Chat~\cite{zheng2023lmsys}, SelfInstruct~\cite{wang2022self}, UltraChat~\cite{ding2023enhancing}, WebGLM-QA~\cite{liu2023webglm}, Unnatural Instructions~\cite{honovich2022unnatural}, WildChat~\cite{zhao2024wildchat}, Wizard-evol-instruct~\cite{xu2023wizardlm}\\
\hline
%3
 Instruction Fine-Tuning & Improvement of Datasets & DialogStudio~\cite{zhang2023dialogstudio}, Dynosaur~\cite{yin2023dynosaur}, Flan-mini~\cite{ghosal2023flacuna}, Flan~\cite{longpre2023flan}, InstructDial~\cite{gupta2022instructdial}, Open-Platypus~\cite{lee2023platypus}OPT-IML~\cite{iyer2022opt}, PromptSource~\cite{bach2022promptsource}, T0~\cite{sanh2022multitask}, UnifiedSKG~\cite{xie2022unifiedskg}, xP3~\cite{muennighoff2022crosslingual}, P3~\cite{sanh2022multitask}, IEPile~\cite{gui2024iepile}  \\
\hline
%4
 Instruction Fine-Tuning & Human \& Improv. Generated  &LIMA-sft~\cite{zhou2024lima}, COIG-CQIA~\cite{bai2024coig} \\
\hline
%5
 Instruction Fine-Tuning & Human \& Model Generated  &InstructGPT-sft~\cite{ouyang2022training}\\
\hline
%6
 Instruction Fine-Tuning & Improv. \& Model Generated &Alpaca-GPT4-data~\cite{peng2023instruction}, Bactrain-X~\cite{li2023bactrian}, Baize~\cite{xu2023baize}, GPT4All~\cite{anand2023gpt4all}, LaMini-LM~\cite{wu2023lamini}, LogiCoT~\cite{liu2023logicot}, LongForm~\cite{koksal2023longform}, OpenOrca~\cite{mukherjee2023orca}, Lithuanian-QA~\cite{nakvosas2024open}, LongWriter~\cite{bai2024longwriter}  \\
\hline
%7
 Instruction Fine-Tuning & Multi-Category &HC3~\cite{guo2023close}, Phoenix-sft~\cite{chen2023phoenix} \\
\hline
%8
 Alignment Fine-Tuning & - & Anthropic-HH-RLHF~\cite{bai2022training},Anthropic-HH-RLHF-2~\cite{ganguli2022red} \\
\hline
%9
 General Purpose & Mulit-Modal &MMRS~\cite{zhang2024earthgpt}, VideoChat2~\cite{li2024mvbench}, InstructDoc~\cite{tanaka2024instructdoc}, ALLaVA~\cite{chen2024allava}  \\
\hline
%10
 Domain Specific & Medical&ChatDoctor~\cite{li2023chatdoctor}, CMtMedQA~\cite{yang2024zhongjing}, DISC-Med~\cite{bao2023disc}, HuatuoGPT~\cite{zhang2023huatuogpt}, MedDialog~\cite{zeng2020meddialog}, Medical Meadow~\cite{han2023medalpaca}, Mol-Instructions~\cite{fang2023mol}  \\
\hline
%11
 Domain Specific & Code &CodeContest~\cite{li2022competition}, ToolAlpaca~\cite{tang2023toolalpacageneralizedtoollearning}, ToolBench ~\cite{qin2023toolllmfacilitatinglargelanguage}  \\
\hline
%12
 Application Specific & Law &DISC-Law~\cite{yue2023disc}  \\
\hline
%13
 Application Specific & Mathematics &Goat~\cite{liu2023goat}, MWP~\cite{lan2022mwptoolkit}, OpenMathInstruct-1~\cite{toshniwal2024openmathinstruct}  \\
\hline
%14
 Application Specific & Education &Educhat~\cite{dan2023educhat} \\
\hline
%15
 Application Specific & Finance & DISC-Fin~\cite{chen2023disc}, AlphaFin~\cite{li2024alphafin} \\
\hline
%16
 Application Specific & Geo science & GeoSignal~\cite{deng2024k2} \\
\hline
%17
 Application Specific & IT & Owl-Instruction~\cite{guo2023owl} \\
\hline
\end{tabular}
    \label{dataset:finetuning}
\end{table*}

\subsection{Datasets for Fine-Tuning}
As mentioned in Section~\ref{sec:fine-tuning}, fine-tuning, as an inseparable part of~\acp{LLM}' development, adapts the pre-trained model to perform specific tasks or to specialize in certain domains~(See Tab.~\ref{dataset:finetuning}).
During fine-tuning, the model is exposed to more focused and often smaller datasets than those used during pre-training phase. Considering this fact, fine-tuning datasets can be categorized based on their applications and use cases.
From another point of view, they can also be categorized based on the types of fine-tuning techniques discussed in more detail in Section~\ref{sec:fine-tuning}. Considering these properties, we categorize fine-tuning datasets into four main groups: instruction-based, alignment-based, domain-specific, and application-specific datasets.  

Instruction-based datasets are utilized in the instruction fine-tuning techniques and contain a series of pairs of instruction and answer texts. Instructions are the input to the model given by the user while answers are the outputs generated by the~\ac{LLM}. Considering the fact that the instructions and outputs in these datasets can be generated by humans, other models, or an improvement of other datasets, they are divided into several groups~\cite{liu2024datasets} as is shown in Tab.~\ref{dataset:finetuning}. It is good to mention that the instructions used in these datasets are general instructions collected for general tasks that are not limited to particular domains or tasks.
On the other hand, alignment datasets such as Anthropic-HH-RLHF~\cite{bai2022training} are utilized for the alignment tuning of the model with human preferences, and are mainly collected manually by human.
Some other datasets, categorized as application-specific datasets in Tab.~\ref{dataset:finetuning}, contain particular contexts for particular applications, allowing the~\ac{LLM} to be optimized in such a way that it shows better performance in the related application. 
\begin{table*}
\caption{List of datasets for the Evaluation stage of~\acp{LLM}.}
\centering
\begin{tabular}{p{2.4cm}p{3.7cm}p{9.8cm}}
\hline
 \textbf{Dataset Domain} & \textbf{Dataset Type} & \textbf{Dataset Names and applied paper}\\
\hline
    %1
 Task Specific &Natural Language Under-
standing &CLUE~\cite{xu-etal-2020-clue}, SuperGLUE~\cite{wang2019superglue}, CUGE~\cite{yao2021cuge}, MCTS~\cite{chong2023mcts}, LeSC~\cite{wu2024can}, CoQA~\cite{reddy2019coqa}, DuoRC~\cite{saha2018duorc}, WiC~\cite{pilehvar2018wic}, Wikitext~\cite{pilehvar2018wic}, LCQMC~\cite{liu2018lcqmc} \\
\hline
%2
Task Specific & Sentence Completion \& Story Cloze &LAMBADA~\cite{paperno-etal-2016-lambada}, ChID~\cite{zheng-etal-2019-chid}, CLOTH~\cite{xie-etal-2018-large}, StoryCloze~\cite{mostafazadeh2016corpus}, AdGen~\cite{shao-etal-2019-long}, CHID-FC~\cite{xu2021fewclue}, HellaSwag~\cite{zellers2019hellaswag} \\
\hline
%3
 Task Specific & World Understanding &ARC~\cite{clark2018think}, OpenBookQA~\cite{mihaylov-etal-2018-suit}, PIQA~\cite{bisk2020piqa}, JEC-QA~\cite{zhong2020jec}, HEAD-QA~\cite{vilares-gomez-rodriguez-2019-head}, WikiQA~\cite{yang2015wikiqa}, ALCUNA~\cite{yin2023alcuna}, KoLA~\cite{yu2023kola}, SocKET~\cite{choi-etal-2023-llms}, LMExamQA~\cite{bai2024benchmarking}\\
\hline
%4
             Task Specific & Contextual Understanding &QuAC ~\cite{choi-etal-2018-quac}, L-EVAL~\cite{an2023eval}, LongBench~\cite{bai-etal-2024-longbench}, ZeroSCROLLS~\cite{shaham-etal-2023-zeroscrolls}, LooGLE~\cite{li2023loogle}, CLongEval~\cite{qiu2024clongeval}, Counting-Stars~\cite{song2024countingstarsmultievidencepositionawarescalable} \\
\hline
 Task Specific & Commonsense Reasoning &CommonsenseQA~\cite{talmor-etal-2019-commonsenseqa}, ECQA~\cite{aggarwal2021explanations}, ReCoRD~\cite{zhang2018record}, SocialIQA~\cite{sap-etal-2019-social}, CREAK~\cite{onoe2021creak} \\

\hline
%6
 Task Specific & Reasoning &Chain-of-Thought Hub~\cite{fu2023chain}, Choice-75~\cite{hou-etal-2024-choice}, STRATEGYQA~\cite{geva2021did}, COPA~\cite{roemmele2011choice}, PROST~\cite{aroca2021prost}, WIQA~\cite{tandon-etal-2019-wiqa} \\
\hline
%7
 Task Specific & Reading Comprehension &BoolQ~\cite{clark2019boolq}, CosmosQA~\cite{huang2019cosmos}, CondaQA~\cite{ravichander2022condaqa},MultiRC~\cite{khashabi2018looking}, RACE~\cite{lai-etal-2017-race}, C3~\cite{sun2020investigating}, ReClor~\cite{yu2020reclor}, DREAM~\cite{sun2019dream}, SQuAD~\cite{rajpurkar-etal-2016-squad}, HOTPOTQA~\cite{yang-etal-2018-hotpotqa}, TriviaQA~\cite{joshi-etal-2017-triviaqa}, Natural Questions~\cite{kwiatkowski2019natural}, CMRC2018~\cite{cui-etal-2019-span}, Adversarial QA~\cite{bartolo2020beat}, Quoref ~\cite{dasigi2019quoref}, DuReader Robust~\cite{tang-etal-2021-dureader}, MS-MARCO~\cite{bajaj2018msmarcohumangenerated}, DROP~\cite{dua-etal-2019-drop}, QASPER~\cite{dasigi-etal-2021-dataset} \\
\hline
%8
         Task Specific & Natural Language Inference \& Logical Reasoning &ANLI~\cite{nie-etal-2020-adversarial}, MNLI-m~\cite{williams-etal-2018-broad}, OCNLI~\cite{hu-etal-2020-ocnli}, CMNLI~\cite{xu-etal-2020-clue}, HANS~\cite{mccoy-etal-2019-right}, WANLI~\cite{liu2022wanli}, MultiNLI~\cite{williams-etal-2018-broad}, SNLI~\cite{bowman-etal-2015-large}, NeuLR~\cite{xu2023large}, LogiQA~\cite{liu2020logiqa} \\
\hline
%9
             Task Specific & Cross-Lingual Understanding &TyDiQA~\cite{clark2020tydi}, MLQA~\cite{lewis-etal-2020-mlqa}, XNLI~\cite{conneau-etal-2018-xnli}, PAWS-X~\cite{yang-etal-2019-paws}, XCOPA~\cite{ponti-etal-2020-xcopa}, XWinograd~\cite{tikhonov-ryabinin-2021-heads}, MLSum~\cite{scialom-etal-2020-mlsum}, XTREME~\cite{hu2020xtreme}, WikiLingua~\cite{ladhak-etal-2020-wikilingua}, MARC~\cite{keung2020multilingual} \\
\hline
%10
             Task Specific & Truthfulness \& Fact Checking &FACTOR~\cite{muhlgay-etal-2024-generating}, FActScore~\cite{min-etal-2023-factscore}, FactualityPrompt~\cite{lee2022factuality}, FreshQA~\cite{vu-etal-2024-freshllms}, HalluQA~\cite{cheng2023evaluating}, HaluEval~\cite{li2023halueval}, TruthfulQA~\cite{lin-etal-2022-truthfulqa}, UHGEval~\cite{liang-etal-2024-uhgeval}, RealTime QA~\cite{kasai2024realtime}, FairEval~\cite{wang-etal-2024-large-language-models-fair}, MultiFC~\cite{augenstein2019multifc} \\
\hline
%11
             Task Specific & Biases and Ethics &ETHOS~\cite{mollas2020ethos}, StereoSet~\cite{nadeem2020stereoset}, BBQ~\cite{parrish-etal-2022-bbq}, Winobias~\cite{zhao-etal-2018-gender}, CrowS-Pairs~\cite{nangia-etal-2020-crows} \\
\hline
%12
           Task Specific & VToxicity &RealToxicityPrompts~\cite{gehman2020realtoxicityprompts}, Safety-Prompts~\cite{sun2023safety}, SafetyBench~\cite{zhang2023safetybench}, TRUSTGPT~\cite{huang2023trustgpt}, HELM~\cite{liang2022holistic} \\
\hline
%13
 Task Specific & Robustness &PromptBench~\cite{zhu2023promptbench} \\
\hline
%14
             Task Specific & Dialogue &Empathetic Dialogues~\cite{dinan2020second},  ConvAI2~\cite{dinan2020second} \\
\hline
%15
             Task Specific & Text Classification, Generation, Translation &RAFT~\cite{alex2021raft}, DART~\cite{nan-etal-2021-dart}, E2E~\cite{novikova2017e2e}, NLLB~\cite{costa2022no} \\
\hline
%16
             Task Specific & Text Summarizing &CNewSum~\cite{wang2021cnewsum}, XL-Sum~\cite{hasan2021xl}, WikiHow~\cite{koupaee2018wikihow}, MediaSum~\cite{zhu2021mediasum} \\
\hline
%17
             Task Specific & Out of Distribution Understanding &GLUE-X\cite{yang-etal-2023-glue}, BOSS~\cite{yuan2023revisiting} \\
\hline
%18
 Task Specific & Elementary Task &LMentry~\cite{efrat2022lmentry} \\
\hline
%19
             Task Specific & Multi-Task &BBH~\cite{suzgun2022challenging}, BIG-bench~\cite{srivastava2022beyond}, decaNLP~\cite{mccann2018natural}, AlignBench~\cite{liu2023alignbench}, CommonGen~\cite{lin2019commongen}, MMLU~\cite{hendrycks2020measuring} \\
\hline
%20
 Application Specific & Medical &cMedQA2~\cite{zhang2018multi}, PsyQA~\cite{sun-etal-2021-psyqa}, WebMedQA~\cite{he2019applying}, PubMedQA~\cite{jin-etal-2019-pubmedqa}, MedNLI~\cite{romanov-shivade-2018-lessons}, CBLUE~\cite{zhang-etal-2022-cblue}, HuaTuo26M~\cite{li2023huatuo}, MultiMedQA~\cite{singhal2023large}, METS-CoV~\cite{zhou2022mets} \\
\hline
%21
             Application Specific & Law &CUAD~\cite{kwiatkowski-etal-2019-natural}, LAiW~\cite{dai2023laiw}, LawBench~\cite{fei2023lawbench}, LegalBench~\cite{guha2024legalbench}, LexGLUE~\cite{chalkidis-etal-2022-lexglue}, LEXTREME~\cite{niklaus2023lextreme}, SCALE~\cite{rasiah2023scale} \\
\hline
%22
             Application Specific & Mathematics &GSM8K~\cite{cobbe2021training}, SVAMP~\cite{patel2021nlp}, ASDiv~\cite{miao2021diverse}, MATH~\cite{hendrycks2021measuring}, Ape210K~\cite{zhao2020ape210k}, Math23K ~\cite{wang2017deep}, MathQA~\cite{amini-etal-2019-mathqa}, AQUA-RAT~\cite{ling-etal-2017-program}, NaturalProofs~\cite{welleck2021naturalproofs},  MGSM~\cite{shi2022language}, MultiArith~\cite{roy2016solving}, AS-Div~\cite{miao2021diverse}, MAWPS~\cite{koncel2016mawps}, TabMWP~\cite{lu2022dynamic}, LILA~\cite{mishra2022lila}, MiniF2F-v1~\cite{zheng2021minif2f} \\
\hline
%23
             Application Specific & Finance &BBF-CFLEB~\cite{lu2023bbt}, FinEval~\cite{zhang2023fineval}, FLUE~\cite{shah-etal-2022-flue}, FinBen~\cite{xie2024finben} \\
\hline
%24
             Application Specific & IT &Owl-Bench~\cite{guo2023owl} \\
\hline
%25
 Application Specific & Geo Science &GeoBench~\cite{10.1145/3616855.3635772} \\
\hline
%26
 Application Specific & Coding &BIRD~\cite{li2024can}, CodeXGLUE~\cite{lu2102codexglue}, DS-1000~\cite{lai2023ds}, HumanEval~\cite{chen2021evaluating}, HumanEvalPack~\cite{muennighoff2023octopack}, MTPB~\cite{nijkamp2022codegen}, ODEX~\cite{wang2022execution}, APPS~\cite{hendrycks2021measuring}, MBPP~\cite{austin2021program}, DuSQL~\cite{wang2020dusql}, CSpider~\cite{min-etal-2019-pilot}, Spider~\cite{yu-etal-2018-spider} \\
\hline
%27
 Application Specific & General Science &Chemical Reactions~\cite{taylor2022galactica}, AminoProbe~\cite{taylor2022galactica}, BioLAMA~\cite{taylor2022galactica}, Galaxy Clusters~\cite{taylor2022galactica}, Mineral Groups~\cite{taylor2022galactica}, SciQ~\cite{welbl2017crowdsourcing} \\
\hline
%28
Domain Specific & Tools &API-Bank~\cite{li-etal-2023-api}, APIBench~\cite{patil2023gorilla}, ToolBench~\cite{xu2023tool}, ToolEyes~\cite{ye2024tooleyes} \\
\hline
%29
          Domain Specific & Mulit-Modal &MVBench~\cite{li2024mvbench},OlympiadBench~\cite{he2024olympiadbench}, MMMU~\cite{yue2024mmmu}, MMT-Bench~\cite{ying2024mmt}, MM-NIAH~\cite{wang2024needle}, MultiTrust~\cite{zhang2024benchmarking}, MMIU~\cite{meng2024mmiu} \\
\hline
\end{tabular}
    \label{dataset:evaluation}
\end{table*}
\subsection{Datasets for Evaluation}
Evaluating~\acp{LLM} involves testing them across a range of tasks to assess their performance. During the evaluation process, not only fundamental tasks such as natural language understanding and natural language generation ones should be evaluated~\cite{Minaee2024LargeLM}, but also safety contents should be assessed properly. In addition to these items, since some models are developed to be used in specific applications, the performance and validity of their responses should also be evaluated within those applications~\cite{liu2024datasets}. 
Due to these reasons, evaluation datasets can be divided into three main groups: task-, application-, and domain-specific ones~(See Tab.~\ref{dataset:evaluation}).
Generally, Task-specific datasets are used to evaluate the ability of the model in~\ac{NLU},~\ac{NLG}, and safety domains. The understanding capacity of the model scan be evaluated by various tasks such as contextual understanding, world understanding, cross-lingual understanding, text classification, natural language inference, question answering, commonsense reasoning, mathematical reasoning, reading comprehension, and problem solving, while the generation capacity of the model can be assessed by different tasks such as summarization, sentence completion, language translation, and dialogue generation. On the other hand, the safety of the models should be assessed from ethical, truthfulness, bias, and toxicity points of view by appropriate datasets.
Moreover, many datasets are gathered to assess the performance of~\acp{LLM} in particular applications such as medical, law, mathematics, finance, and IT fields. These datasets mainly focus on specific properties of the application in addition to evaluating the~\ac{NLU},~\ac{NLG} ability of the model. In this regard, we first categorize the evaluation datasets based on their use cases and applications and then based on their tasks, as is shown in Tab.~\ref{dataset:evaluation}. 
\section{ChatGPT and~\ac{LLM} Applications:}\label{application}
In this section, we will have an overview of ChatGPT, a cutting-edge conversational language model built on~\ac{GPT} applications and finally, we will summarize and present a list of these applications in Table~\ref{LLM_App}. To this end, the survey paper~\cite{NAZIR2023100022} highlights ChatGPT's numerous applications in various fields; the paper also addresses its drawbacks, difficulties, and possible solutions.
\subsection{Anomaly Detection}
 ~\acp{LLM} and ChatGPT significantly enhance anomaly detection efforts by providing advanced~\ac{NLP} capabilities, real-time analysis, and intelligent insights. Their ability to process and understand complex data patterns makes them valuable tools for detecting and responding to anomalies across various domains. Integrating~\acp{LLM} into anomaly detection frameworks can lead to more accurate, efficient, and user-friendly monitoring systems. \\
 The research \cite{shao2022log} proposes a log anomaly detection method, called Prog-~\ac{BERT}-\ac{LSTM}, to quickly and accurately detect system faults from log text data. It does this by using a network that utilizes the BERT model as the text vectorization module and designing the sequence feature learning module based on~\ac{LSTM} to prevent the loss of sequence features caused by the gradient disappearing during the calculation process and to further obtain the semantics. To aggregate the text semantic vector and sequence feature vector, the progressive masking technique is applied. \\
 This article \cite{balasubramanian2023transformer} presents an innovative approach to using anomaly detection features in chatbots to improve their performance. Their chatbot becomes increasingly adept at spotting anomalies by identifying and extracting odd patterns and deviations from logs using sophisticated~\ac{GPT}-3 models and rule-based reasoning. The authors outline the design and process of their anomaly detection system and demonstrate its practical application. Their chatbot, which combines domain knowledge and artificial intelligence, redefines the bar for interactive, anomaly-aware conversational agents. \\
 Manually analyzing the growing amount of log data generated by software-intensive systems is not feasible. Many domains have seen encouraging outcomes via ChatGPT. Studies on the use of ChatGPT for log-based anomaly detection are still lacking, nevertheless. The authors of \cite{qi2023loggpt} suggested LogGPT, a log-based anomaly detection system based on ChatGPT, to close this gap. LogGPT seeks to investigate the transferability of knowledge from large-scale corpora to log-based anomaly detection by utilizing~\ac{ChatGPT}'s language interpretation skills.  LogGPT exhibits strong interpretability and yields encouraging findings. \\
 Using~\ac{GPT}-3 language models, the authors of \cite{mannam2023optimizing} suggest a unique method for log anomaly identification. They turn log data into a language model that can spot odd patterns and anomalies by using the word embedding and tokenizer features of~\ac{GPT}-3. Their suggested approach can be combined with software release management procedures to enhance quality control and automatically identify abnormalities.  \\
 Through extensive data analysis, the study \cite{10452613} seeks to improve pretrained~\acp{LLM}' ability to identify abnormalities and vulnerabilities. We have created a vulnerability detector based on the~\ac{ChatGPT} 3.5 model that performs ordinal vulnerability assessment. \\
The study \cite{lai2023intrusion} investigates how~\ac{LLM}—specifically,~\acp{BERT}—can be applied to intrusion detection systems. With the emergence of sophisticated cyber threats, there is an increasing need to design sophisticated intrusion detection systems (~\ac{IDS}). The study suggests a unique paradigm for identifying abnormal activity and extracting valuable information from network data using~\ac{BERT}. Through network data transformation into a format compatible with natural language, the model efficiently identifies patterns that conventional~\ac{IDS} often misses.  \\ 
The authors of \cite{khediri2024enhancing} offer a novel solution to this problem in accordance with~\ac{IDS} by fusing SHAP (SHapley Additive exPlanations) values with~\acp{LLM}. This technique uses the CICIDS2017 dataset to show how the combination makes it easier to generate human-understandable explanations for anomalies that are found, with the goal of improving transparency and trust in~\ac{IDS}. Significant aspects indicated by SHAP values are articulated by the~\ac{LLM} effectively, providing logical answers for important determinants of model results. \\
As the benchmark for~\ac{AI} quality analysis, a refined~\ac{GPT}-based sentiment analysis model is initially built and examined in \cite{ouyang2023quality}. Subsequently, the data adequacy quality analysis is carried out, which involves using the content-based approach to produce reasonable adversarial review comments as the incorrectly-annotated data and creating approaches based on surprise adequacy (SA) to identify these abnormal data. \\
By integrating~\acp{LLM} with computer vision systems, the technology can describe images, identify objects, and even detect anomalies in visual data. For example, the paper \cite{wang2024visiongpt} explores the use of~\acp{LLM} for zero-shot anomaly detection to enhance secure visual navigation. The proposed framework detects anomalies in camera frames, provides concise audio descriptions of these anomalies, and supports safe navigation using the Yolo-World object detection model and customized prompts. This approach leverages the strengths of~\acp{LLM} and open-vocabulary object detection to dynamically adapt to changing scenarios, improving over traditional visual navigation methods. \\
Large Vision-Language Models (LVLMs) excel at identifying common objects due to their extensive training datasets. However, they struggle with specialized domain knowledge and localized object details, making them less effective for Industrial Anomaly Detection (IAD). Current IAD methods typically produce anomaly scores and require manual threshold setting to distinguish between normal and abnormal samples, limiting their practical application. Regarding this, the authors in \cite{gu2024anomalygpt} propose AnomalyGPT, an innovative LVLM-based approach to IAD. This method involves generating detailed written descriptions and simulating anomalous images to create training data. An image decoder provides fine-grained semantic information, while a prompt learner fine-tunes the LVLM using prompt embeddings. AnomalyGPT can directly identify the locations and presence of anomalies, eliminating the need for manual threshold adjustments.
\subsubsection{Fake Detection}
Despite advancements in technology, there is a notable research gap in using advanced technologies like LLMs, such as~\ac{ChatGPT}-3.5 and Bard, for fake news detection.~\ac{ChatGPT} and~\acp{LLM} can aid in fake news detection by analyzing textual data for patterns and inconsistencies, assessing news credibility, and providing real-time analysis by cross-referencing reliable sources. The study \cite{teo2024integrating} aims to bridge this gap by exploring and evaluating the effectiveness of combining traditional machine learning techniques with~\ac{ChatGPT}-3.5 and~\ac{LLM} judgments for identifying fake news. The superior performance of this model is attributed to the inclusion of~\ac{ChatGPT}-3.5's authenticity assessments, which underscore the importance of nuanced linguistic patterns in distinguishing between real and fake news. \\
In the study \cite{huang2023fake},~\ac{ChatGPT} is utilized to neutralize text, enabling the authors to compare the original and neutralized versions. This comparison reveals differences, such as varying rates of sentiment word usage, which can help identify fake news. They observed that fake news generally contains more sentimental, particularly negative, language than real news. Leveraging this insight can improve fake news detection outcomes. \\
The coherence between an image and two captions, as well as between two captions, is assessed by the authors in \cite{wu2023cheap} using the~\ac{LLM} structures. Initially, the technique uses an IoU value to assess the image-caption coherence. Next, it estimates a similarity vector using S-BERT and BERT-large models. Furthermore, based on a collection of thoughtfully created features, they employ the~\ac{GPT}-3.5 model to create a discriminative vector that depicts the semantic relationship between the captions. \\
Given the rise of fake news on social media, the authors in \cite{koru2024detection} aim to identify accurate information on Twitter. They collected true news from mainstream newspaper Twitter accounts and fake news from Twitter verification platforms, preprocessing the data using the Zemberek~\ac{NLP} tool to create the TR-FaRe-News dataset. They applied ensemble approaches and vectorization techniques like BoW, TF-IDF, and Word2Vec to this dataset for fake news detection. Furthermore, they adjusted a pre-trained BERT deep learning model and explored different model iterations, incorporating CNN layers and Bi-LSTM with frozen and unfrozen parameter techniques, to enhance detection accuracy. \\
The authors of \cite{ayoobi2023looming} introduce an innovative method for swiftly identifying fake and~\ac{LLM}-generated profiles on LinkedIn before connections are made. Early detection is crucial for maintaining platform integrity by preventing imposters from accessing sensitive user data and building credibility for future phishing and scamming activities. They utilize textual data from LinkedIn profiles and propose the Section and Subsection Tag Embedding (SSTE) technique, which enhances the discriminative properties of profile data. This technique effectively differentiates between authentic profiles and those created by imposters, whether manually or through automated LLMs. \\ 
\subsection{Conversation Development and Chatbot}
One significant advancement in~\ac{AI} is the emergence of several chatbots that can simulate human-like speech. Since these "fluid conversationalists" have already passed multiple human-designed exams and have been employed by students in winter and autumn 2022 exams, their appearance has piqued the interest of many people, particularly professors. In the following,~\ac{ChatGPT}, the most visible member of this family, is used to dispel some common misconceptions about these programs \cite{henno2023we}.~\ac{ChatGPT} facilitates the growth of conversations by offering responses that are contextually appropriate, guaranteeing seamless and captivating exchanges. It can mimic various conversational situations, which aids in the improvement of dialogue systems by developers. Furthermore,~\ac{ChatGPT} provides instantaneous feedback and recommendations for enhancing the coherence, flow, and engagement of conversations.\\
The study \cite{sudharson2023abstractive} investigates the main elements, difficulties, and uses of such systems as they relate to abstractive summarization and large-language modeling techniques for question-answering. It highlights the application of large language models for conversational~\ac{AI} and abstractive summarization, such as~\ac{GPT}-3. Several model variations, including the Flan T5 and LaMini Flan T5, are used in this article. They also go into moral issues and the appropriate application of these technologies, stressing openness, reducing bias, and protecting user privacy. \\
In order to comprehend the significance of library versions in discussions pertaining to code, the authors in \cite{raj2024role} examine DevGPT, a dataset comprising over 4,000 Developer-~\ac{ChatGPT} interactions. They measure the frequency with which library version restrictions are brought up in discussions about code and the instances in which~\ac{ChatGPT} suggests installing particular libraries. To ensure higher-quality responses, version limits are typically suggested by users during talks rather than being set by~\ac{ChatGPT}. \\
In order to gain a better understanding of how developers pinpoint regions in code that require modification and how~\ac{ChatGPT} meets those needs, the authors of \cite{alomar2024refactor} examine discussions between developers and~\ac{ChatGPT} pertaining to refactoring. Their method is based on looking into developers' clear refactoring intentions and text mining refactoring-related conversations from 17,913~\ac{ChatGPT} prompts and responses. \\
In order to investigate how software developers engage with~\ac{ChatGPT}, a well-known LLM, the study \cite{xiao2024devgpt} presents DevGPT, a dataset that has been carefully selected. The dataset contains 29,778~\ac{ChatGPT} questions and responses, 19,106 code snippets, and is connected to related software development artifacts like discussions, pull requests, issues, source code, and Hacker News threads. DevGPT makes it possible to investigate developer inquiries, the efficiency of~\ac{ChatGPT} in generating code and resolving issues, and the more general applications of artificial intelligence in programming. \\
The authors \cite{mohamed2024chatting} share the findings of an empirical study that found trends in the conversations developers have with~\acp{LLM} in order to provide a study aimed at understanding how developers interact with and use LLMs. In their talks with LLMs, they discovered a total of 19 topics that explained the developers' goals. \\
In \cite{pereira2023fuzzy}, the authors suggest combining BERT and RoBERTa to perform Emotion Recognition in Conversations (ERC) in order to provide classifiers based on Large Language Models that are easier to understand and more straightforward. In order to produce contextually embedded utterance representations, they suggest feeding the utterances along with their prior convergenceal turns to a pre-trained RoBERTa. These representations would then be fed to an updated Fuzzy Fingerprint classification module.  \\
The goal of the study \cite{eagalapati2024enriching} is to strengthen~\ac{ChatGPT}'s ability to comprehend and react to visual input by exploring the integration of picture captioning approaches. The paper delves deeply into the use of deep learning models, including MobileNet, ResNet50, LSTM, DenseNet121, and MobileNetv2, in the context of image captioning. In particular, a thorough analysis is carried out on a Recurrent Neural Network that uses LSTM as a decoder and a Convolutional Neural Network that uses ResNet as an encoder. These fusions use image characteristics and words to create accurate and insightful descriptions of visual material.  \\
It might be difficult for non-native speakers to become fluent and locate chances for language practice. The developers of \cite{ukirde2023english} offer a fresh solution to these problems by creating a language matching algorithm and a fluency detection model.
The fluency detection model evaluates non-native speakers' fluency using machine learning approaches.  A linguistic matching algorithm that seeks to pair non-native speakers with native speakers for language practice is another addition to the fluency identification model. \\
The authors of \cite{song2024innovative} examine a~\ac{ChatGPT}-based Japanese speech dialogue practice system that provides personalized language instruction. They replicate Japanese speech with standard pronunciation, accents, and varied speech rates to train dialogue generation models. Using voice synthesis, they simulate speech traits of different speakers and implement virtual dialogues. The system integrates speech recognition and a pronunciation assessment tool to provide immediate feedback to students. They aim to enhance~\ac{ChatGPT}'s understanding and production of speech dialogues in everyday, affective, and contextual conversations, offering students a tailored learning experience through real-time interactions and feedback. \\
Advanced methods of holding free-flowing conversations or debates between humans and~\acp{LLM} yield far more fascinating outcomes, showcasing problematic rhetorical behaviors like evasion, circular arguments, self-contradictions, topic changes, inconsistent positions, and the combination of passive aggression and attempts to appease the human disputant. During a~\ac{ChatGPT} debate session about translating Japanese song lyrics, the writers of \cite{selitskiy2024yet} share their unique observation of this type of behavior. \\
In \cite{liu2024transformer}, the authors present an Automatic conversation model (ACC) based on the Transformer BERT ensemble model for English settings, with the goal of enhancing the current automatic conversation model.  The Transformer BERT model's deep self attention mechanism and pre-trained contextual comprehension abilities allow it to better capture the relationship between texts and phrases, which enhances the dialogue model's performance. In terms of accuracy and efficiency, the Transformer BERT integrated model-based ACC automatic conversation model performs better than other benchmark models. \\
The software development community's quick adoption of~\ac{ChatGPT} has created new opportunities to investigate the qualitative and quantitative effects of the platform on Developer-~\ac{ChatGPT} talks. In order to conduct a comprehensive analysis, the writers of \cite{deo2024analyzing} dig into a rich dataset from GitHub and Hacker News. Characterizing the nature of these interactions and assessing~\ac{ChatGPT}'s use in refactoring are among their goals. They use a combination of exploratory data analysis and data annotation to accomplish these goals, extracting important information with the help of keyword filters. \\
\subsection{Healthcare}
\ac{ChatGPT} and~\acp{LLM} assist in healthcare by providing accurate medical information, supporting diagnosis through symptom analysis, and aiding in patient communication. They can analyze vast amounts of medical data to identify patterns and assist in personalized treatment plans. Additionally, these models help streamline administrative tasks, enhancing overall healthcare efficiency. To provide an understanding of what~\ac{AI}-based systems in the healthcare industry look like, medical images will be interpreted utilizing Deep Learning, Generative~\ac{AI}-based LLMs, and~\ac{NLP} for Healthcare Records \cite{sathe2023comprehensive}. \\
More applications of~\ac{LLM} in biological and healthcare, education, social, business, and agricultural is then covered by the authors in \cite{raiaan2024review}. \\
Effective management of Parkinson's disease (PD) requires vigilant monitoring, particularly as the condition progresses and impacts patients' and families' quality of life. The creators of AutoHealth, introduced in \cite{shankar2024deep}, offer an advanced Internet of Medical Things (IoMT) solution leveraging smartwatches with sophisticated features and biosensors. AutoHealth employs vector-based learning~\ac{AI} models to continuously track the movement patterns of PD patients, enabling precise and personalized early detection, ongoing monitoring, and rehabilitation management. Additionally, an~\ac{AI} chatbot enhances user engagement by responding to patients' text and voice queries, providing personalized advice in real-time. \\
The healthcare sector requires~\acp{LLM} that adhere to regulations, ensure privacy, and offer robust security \cite{kuzlu2023rise}. In \cite{rahman2023survey}, the authors explore various multimodal~\acp{LLM} and the associated security and privacy issues that must be addressed. The study highlights security and privacy vulnerabilities identified by researchers and regulatory bodies and discusses protective measures such as federated learning, differential privacy, and monitoring~\ac{LLM} procedures to mitigate these risks. \\
The authors of \cite{alamchatgpt} explore the application of~\ac{ChatGPT} in various medical contexts, demonstrating its potential to write clinical vignettes, patient discharge summaries, and radiology reports, as well as identify cardiovascular disease prevention strategies. They argue that~\ac{ChatGPT} can revolutionize medical education, empower patients with personalized healthcare information, and streamline several healthcare tasks by leveraging its language generation and processing capabilities. This includes digitizing clinical notes and enhancing diagnostic accuracy. \\
Relying on trust and the threat of hostile assaults highlight the necessity of deploying robust security measures with~\ac{ChatGPT} and~\ac{AI}, paying special attention to logical and morally sound products \cite{mohammed2024chatgpt}. \\
\subsubsection{Diagnostic Assistance: Providing differential diagnoses}
\ac{ChatGPT}, and ~\ac{LLM}, are developed to simulate human conversation and respond to questions from various fields, including healthcare. Trained on vast web content, it provides text-based answers and serves as a valuable information source. However, it is not a substitute for a healthcare provider’s expertise and personalized care. Instead,~\ac{ChatGPT} aims to enhance information accessibility and complement healthcare services \cite{panagoulias2023evaluating}. While it is useful for general principles and educational resources, it is essential to consult licensed healthcare providers for medical advice or diagnosis, as they can consider individual circumstances and interpret complex medical data \cite{mosaiyebzadeh2023empowering}. \\
\subsubsection{Patient Communication}
~\ac{LLM} contribution in patient communications includes medical advice, care instructions, medication reminders, and appointment scheduling.
The domains of medication advising and adverse drug response prediction present great potential for the rapidly developing field of LLMs. In spite of this, current~\acp{LLM} find it difficult to handle complex polypharmacy situations. The study \cite{dou2023shennonggpt} introduces ShennongGPT, an advanced~\ac{LLM} designed for reliable medication advising and predicting adverse drug responses. ShennongGPT utilizes a two-stage training process: initially, it learns fundamental drug interaction knowledge from condensed drug databases; subsequently, it simulates human-like decision-making using real patient data to enhance the relevance and applicability of its recommendations. This dual strategy enables ShennongGPT to better predict potential adverse drug interactions and provide personalized medication advice, significantly improving pharmaceutical safety and the overall quality of healthcare services. \\
The difficulties in implementing~\acp{LLM} in medical chatbots for chronic disease self-management are covered in the paper \cite{montagna2024llm}. As a result, the authors describe an architecture that was created especially to address problems with privacy, clinical trials, and reliability. A locally deployed~\ac{LLM} using open-source models and a filtering system for sensitive data with an external~\ac{LLM} are contrasted as ways to prevent data leakage.  \\
Multi-agent framework with an~\ac{LLM} foundation that is intended to automate certain administrative tasks in therapeutic environments. These~\ac{LLM} agents work together to deconstruct tasks, interpret instructions, and carry out a workflow's series of activities. Their capabilities encompass not just database-level documentation execution but also direct web-based electronic medical record (EMR) platform operation \cite{gebreab2024llm}. \\
The authors of \cite{malkiel2024segllm} present a brand-new technique for accurate and efficient call segmentation and topic extraction termed SegLLM. There are offline and online phases in SegLLM. Using a LLM, the offline phase generates a distribution of synthetic sentences for each topic after it is applied to a specified list of topics. The similarity between the transcripted conversation and the topic anchors identified in the offline phase is scored in the online phase, which is applied to each call individually. \\
The authors of \cite{natarajan2024enhancing} utilize the advanced Llama2 model to develop a medical chatbot that emphasizes instance training through the addition and modification of metadata, enabling it to continually update and stay current with medical advancements. This chatbot can respond to inquiries and extract information from a comprehensive meta-dataset, establishing itself as a reliable source of information. Its extensive understanding of medical terminology allows it to provide accurate and timely responses to medical questions, enhancing its credibility and usefulness as a medical resource. \\
The possible effects of~\acp{LLM} on pharmacy education and practice are examined by the writers in \cite{angel2023clinical}. They use a sample of 137 multiple-choice questions from the NAPLEX test to assess the LLMs. Through its individual user interfaces,~\ac{GPT}-3,~\ac{GPT}-4, and Bard are given these questions. The answers produced by the~\acp{LLM} are then compared to the answer key. \\
\subsection{Code Generation and Completion}
By comprehending natural language prompts and producing context-aware, executable code snippets,~\acp{LLM} such as~\ac{ChatGPT} are used for code creation and completion. They help with things like restructuring code, creating boilerplate in various programming languages, and autocompleting methods. The authors of~\cite{gong2024ast} provide Abstract Syntax Tree (AST)-T5, a novel pretraining paradigm that makes use of the AST to enhance code generation, translation, and comprehension.
While their AST-Aware Span Corruption aim provides the paradigm with the means to reassemble various code structures, their AST-Aware Segmentation uses adaptive programming to maintain code structure. The authors in~\cite{li2022competition} present AlphaCode, a code generation system capable of coming up with original answers to these issues that call for more complex thinking.\\
Lai~\textit{et al.} in~\cite{lai2023ds}
 introduce DS-1000, a benchmark for code creation that covers seven Python libraries and includes 1,000 data science tasks taken from StackOverflow. It has a wide range of useful applications, a very accurate automatic evaluation system with a 1.8\% error rate, and problem-modification protections against memorizing.\\
ODEX is an open-domain NL-to-Python dataset that supports multiple languages and contains 945 NL-Code pairs from 79 libraries and 1,707 test cases. It shows behavioral differences between the best code models, with CODEGEN getting better with scale and CODEX performing better overall~\cite{wang2022execution}.
\subsubsection{Software Development}
\ac{ChatGPT} enhances software development by automating processes like code creation, documentation generation, and debugging, improving efficiency and productivity. The rapid uptake of~\ac{ChatGPT} by the software development community has opened up new avenues for generating code snippets, suggesting completions, debugging on the Developer-~\ac{ChatGPT} discussions~\cite{deo2024analyzing}.
\subsection{Content Creation}
The authors in~\cite{liu2022wanli}
suggest a collaborative approach between humans and~\ac{AI} for creating datasets that combines human assessment with the creative capabilities of~\ac{GPT}-3. They find difficult reasoning patterns using MultiNLI as a foundation, produce analogous examples, and then improve them with human input. 
 \subsubsection{Marketing and Social}
 Although social media data~\cite{Gokaslan2019OpenWeb}
is essential for scientific research, it necessitates a high level of processing capacity and technical know-how. Despite being easier to use than Facebook or Twitter, Reddit, a well-known research network, nonetheless poses technical difficulties for data collecting. These days, datasets are regarded as important research tools. A thorough, continuously updated history of Reddit data since its birth is provided by the Pushshift Reddit dataset. Researchers can save time on data preparation and collecting by using its search, aggregation, and analysis features~\cite{baumgartner2020pushshift}. In order to protect user information and platform integrity, the authors~\cite{ayoobi2023looming} provide a technique to identify phony and~\ac{LLM}-generated LinkedIn profiles during registration. With little training data, they provide the Section and Subsection Tag Embedding (SSTE) approach, which achieves approximately 90\% accuracy for~\ac{LLM}-generated profiles and approximately 95\% accuracy for phony profiles. For this study, they also make available the first sizable LinkedIn dataset.
\subsection{Education and Tutoring}
ChatGPT improves software development by automating processes like code creation, documentation generation, and debugging, improving efficiency and productivity. It helps for generating study materials, lessonplans, and, articles
 There is a revolution taking place as machines and gadgets are being created to comprehend, assess, and analyze languages employing LLMs~\cite{raiaan2024review}.\\
 A free and open-source~\ac{LLM}-based chatbot for education, EduChat1 provides teachers, students, and parents with individualized, equitable, and encouraging conversations. It pre-trains on educational data and fine-tunes using system prompts to improve features like Q\&A, essay assessment, Socratic teaching, and emotional support~\cite{dan2023educhat}.\\
 The authors of~\cite{alamchatgpt}
explore the application of~\ac{ChatGPT}
in various medical contexts, demonstrating its potential
to write clinical vignettes, patient discharge summaries,
and radiology reports, as well as identify cardiovascular
disease prevention strategies.\\
The possible effects of~\acp{LLM} on pharmacy education
and practice are examined by the writers in~\cite{angel2023clinical}. To evaluate the LLMs, they utilize a sample of 137 multiple-choice questions taken from the NAPLEX exam. Through each user.These questions are presented to Bard,~\ac{GPT}-3,~\ac{GPT}-4, and interfaces.
The LLMs' responses are then contrasted with the answer key.
\subsubsection{Language Learning}
Using language learning, one can easily perform grammar correction, vocabulary building, conversations. OpenAI and Google~\ac{AI} developed language processing based on contextual word embedding, GPT~\cite{radford2018improving}, and BERT~\cite{devlin2018bert} using transformers.
In addition to creating robust test suites that enable the investigation and comparison of language-specific approaches, the Cross-Language Evaluation Forum has promoted research in text retrieval strategies for many European languages~\cite{mcnamee2004character} and also in text similarity detection~\cite{wang-etal-2018-glue}. 
 Chinese language~\cite{xu-etal-2020-clue} and~\cite{yao2021cuge}
and Chinese biomedical language~\cite{zhang-etal-2022-cblue}
processing and understanding are also explored separately using~\acp{LLM} .
 \subsubsection{Academic Research and Summarizing research papers} 
Academic research involves in-depth investigation of subjects for new knowledge.~\ac{ChatGPT} simplifies complex academic material into understandable summaries, aiding in the comprehension of key ideas and conclusions. MAP-Neo is a 7B parameter bilingual~\ac{LLM} that is totally open-source and was trained using 4.5 trillion high-quality tokens~\cite{zhang2024map}. Its pre-training corpus, data pipeline, checkpoints, and training framework are all fully transparent, and its performance is comparable to that of the most advanced models. This paradigm encourages more innovation and promotes transparency in~\ac{LLM} research.\\
A collection of 81.1 million scholarly works from many fields, S2ORC includes structured full text for 8.1 million open-access papers, metadata, abstracts, and resolved references. It has inline annotations for figures, tables, and citations that are connected to their original sources. Research in academic text mining is supported by S2ORC, the largest public collection of machine-readable academic text that aggregates papers from multiple publishers~\cite{lo-etal-2020-s2orc}.
\subsubsection{Education: Assisting students}
Education is crucial for professional and personal growth, and programs like~\ac{ChatGPT} offer personalized assistance for studying, understanding complex subjects, and enhancing learning results. The study~\cite{ibrahim2024does} explores how~\ac{ChatGPT} affects undergraduate students in Malaysia and finds that it is widely used for learning and assignments. Although it has a good impact on education worldwide, it also emphasizes how important it is to keep an eye on and encourage communication in order to preserve students' inventiveness and creativity.\\
160 students and 80 teachers participated in~\cite{ergene2024ai}, which assessed~\ac{ChatGPT} versions (\ac{GPT}-4o, \ac{GPT}-4, \ac{GPT}-3.5), MathGPT, and Gemini on 390 interactive math questions. Gemini had the lowest success rate, while~\ac{GPT}-4o marginally beat~\ac{GPT}-4, followed by MathGPT and~\ac{GPT}-3.5. Although they acknowledged its limits, educators and students cited~\ac{ChatGPT}'s clear explanations, prompt response, and learning assistance as advantages. All things considered,~\ac{ChatGPT} is recommended as a useful resource for solitary math education.\\
With an emphasis on its uses, potentials, difficulties, and future possibilities, the authors in~\cite{MEMARIAN2023100022}
study 63 papers using~\ac{ChatGPT} in education. Benefits including cognitive support, teaching activities, assessments, and individualized learning have been found, despite the fact that much research is non-empirical. Plagiarism, abuse, responsibility, and privacy issues are among the difficulties. While examining~\ac{ChatGPT}'s educational potential, the article places a strong emphasis on maintaining academic integrity and student learning.\\
With features like idea generation, language acquisition, and individualized support,~\ac{ChatGPT}—which OpenAI introduced in November 2022—offers substantial educational advantages. But issues like excessive dependence and a lack of creativity need to be addressed. In order to improve learning, educators should help students use~\ac{ChatGPT} responsibly by fusing its results with their own observations~\cite{10442033}.\\
By asking 102 high school and college students about their familiarity, usage, effectiveness, and attitudes toward telling teachers about~\ac{ChatGPT}, the study~\cite{khoso2023use} investigates the educational impact of the app. The study stresses the significance of carefully using~\ac{AI} to foster critical thinking and intellectual development, while also outlining advantages and difficulties.
\subsubsection{Writing Assistance}
Writing aids enhance content quality, organization, and clarity by providing advice, correcting drafts, and assisting users with the writing process, such as ChatGPT. According to~\cite{najah2024improving}, including~\ac{ChatGPT} into higher education greatly improves students' descriptive writing abilities, as seen by the average scores increasing by 33.6\% (from 64 to 85.5).~\ac{ChatGPT} raised students' passion for writing while enhancing sensory detail, coherence, and clarity. Its potential to develop academic writing abilities is highlighted by the fact that its efficacy depends on systematic instruction and careful integration with conventional approaches.\\
In~\cite{hidayatullah2024evaluating}, the usage of~\ac{ChatGPT} by students to improve their English writing abilities while avoiding plagiarism is investigated. Through qualitative techniques such as exam analysis, interviews, and observation, it seeks to guarantee that students incorporate~\ac{ChatGPT} into their education in a responsible and moral manner.\\
Using descriptive qualitative analysis, this study investigates~\ac{ChatGPT}'s English essay writing skills~\cite{fitria2023artificial}. Essays on a range of subjects are produced using~\ac{ChatGPT} while preserving proper tenses, sentence diversity, and structure. Although efficient, more investigation is required to assess the outputs' grammatical accuracy.\\
The advantages and disadvantages of~\ac{ChatGPT} in improving high school writing abilities are investigated in~\cite{sain2025benefits}. Although it facilitates article idea production, issues include data accuracy, moral dilemmas, and effects on critical thinking.\\
\subsubsection{General Research}
For general research,~\ac{ChatGPT} is summarizing information and maintaining knowledge bases. A radiology report's "Impression" section provides a summary of the findings, which is frequently prone to inaccuracies. Although automatic impression generation (AIG) shows potential with deep learning models like BERT, these models struggle with generalization and require big datasets. With the help of~\acp{LLM} like~\ac{ChatGPT}, "ImpressionGPT" enhances AIG through iterative optimization and dynamic prompts, producing improved outcomes without the need for more training data. It offers a fresh strategy for using~\acp{LLM} in specialized sectors and performs better on medical datasets than current approaches~\cite{ma2024iterative}.\\
Several approaches to text summarization have been investigated, such as extractive and abstractive methods~\cite{yang2023exploring}. Less research has been done on how well~\ac{LLM} function in aspect-based summarization, though.~\ac{ChatGPT}'s performance on four benchmark datasets is comparable to that of conventional fine-tuning techniques, according to this evaluation.\\
Numerous techniques, such as extractive and abstractive methods, have been proposed for text summarization.~\ac{LLM}-generated news summaries are comparable to human ones, according to recent studies. Using four benchmark datasets—Reddit posts, news articles, meetings, and stories—this study assesses ChatGPT and finds that its performance is on par with more conventional fine-tuning techniques based on Rouge ratings~\cite{zhang2023summit}.\\
The authors of~\cite{tariq2023assessing} use~\ac{ChatGPT} and assess its performance on a variety of textual tasks involving dietary supplement information. They discover that~\ac{ChatGPT} did a respectable job of extracting relations, simplifying, and summarizing texts. However, human evaluators reveal that~\ac{ChatGPT}'s output loses some relevant information in roughly one-third of words.\\
Using four human evaluation techniques on five datasets, the authors of~\cite{gao2023human} investigate~\ac{ChatGPT}'s capacity to carry out human-like summary evaluation. Outperforming certain automatic measures,~\ac{ChatGPT} successfully finished annotations utilizing pyramid, pairwise comparison, Likert scale rating, and binary factuality evaluation. Additionally, the study looks at created explanations and invalid responses, compares~\ac{ChatGPT}'s performance with human evaluation, and assesses the effects of various prompts.
\subsection{Language Modeling and Machine Translation}
\ac{ChatGPT} utilizes advanced neural network techniques for human-like writing, improving language modeling and machine translation. It enhances coherence and fluency in language modeling and reliably translates text between languages, improving communication.
\subsubsection{\ac{NLP} and Improving translation}
By treating monolingual data as additional parallel data and integrating it with automatic back-translation, the study,   ~\cite{sennrich-etal-2016-improving} enhances neural machine translation (NMT). This method sets new state-of-the-art findings by greatly improving performance on the IWSLT 14 Turkish-English (+2.1-3.4 BLEU) and WMT 15 English-German (+2.8-3.7 BLEU) tasks. Performance on the IWSLT 15 English-German task is also improved by fine-tuning with in-domain data. Similar study~\cite{sennrich-etal-2016-neural} shows that subword models outperform a back-off dictionary baseline, achieving a 1.1 BLEU improvement for English-German and 1.3 BLEU for English-Russian on the WMT 15 tasks.\\
Google's Neural Machine Translation (GNMT) system's primary contribution is its deep LSTM-based architecture, which handles unusual words with sub-word units (also known as "wordpieces") and attention techniques~\cite{zhang2016google}. Low-precision arithmetic speeds up translation, and a focused beam search increases quality. When compared to Google's phrase-based approach, GNMT reduces translation errors by 60\% and obtains competitive performance on WMT'14 English-to-French and English-to-German.\\
Cross attention, or global attention~\cite{gheini-etal-2021-cross} is
mainly used in encoder-decoder architectures to find the dependencies between different positions of encoder and decoder
sequences. \\
Local attention~\cite{luong-etal-2015-effective} mainly focuses on finding the
dependencies of subset of sequences in the decoder part by
avoiding the expensive computations.\\
Edit-distance~\cite{przybocki2006edit} as the most common
character-based metrics determines the minimum number of
single-character adjustments (insertions, deletions, or replacements) needed to transform a word or text string into another
that can be helpful when assessing spelling corrections or tasks
in which fine-grained text accuracy is crucial.\\
Word-based metrics evaluate the quality of the generated text at the word level, providing insights into the
model’s linguistic capabilities. Bilingual Evaluation Understudy (BLEU), Recall-Oriented Understudy for Gisting Evaluation (ROUGE), and Metric for Evaluation of Translation
with Explicit Ordering (METEOR) are some the most common
word-based metrics. BLEU~\cite{papineni2002bleu} compares the LLM’s output
to annotated ground truths by measuring the precision of ngrams in the generated text against one or more reference texts\\
\subsection{Customer Support and Answering queries}
\ac{ChatGPT} improves customer service by answering questions quickly and accurately and by being available around-the-clock. By streamlining communication, it increases customer happiness and solves frequent problems more quickly.
The study~\cite{zierock2023leveraging} examines how prompting might enhance~\ac{AI}-powered customer support systems, with a particular emphasis on~\ac{ChatGPT} and Midjourney. Model performance, interpretability, and personalization are all improved by prompting. In order to improve consumer interactions in these systems, it emphasizes timely design and execution.\\
Based on transformer architecture,~\ac{ChatGPT} improves patient and customer care by facilitating better communication, managing numerous requests, and offering round-the-clock assistance. It enhances patient experiences in healthcare by bridging linguistic barriers. Even though it is revolutionizing the service industries, there are still issues with guaranteeing accurate and current information~\cite{haleem2024exploring}.\\
The authors in~\cite{limna2023role} investigate how~\ac{ChatGPT} affects customer service in the hospitality sector in Krabi, Thailand.~\ac{ChatGPT} enhances employee abilities, overcomes language barriers, provides advice, and increases productivity, according to 15 stakeholders interviewed for the study.\\
The study~\cite{ezenkwu2023towards}
investigates how companies might use~\ac{ChatGPT} and other~\ac{AI} technologies to improve customer engagement and loyalty. It talks about how professional system development and timely engineering are required to customize~\ac{ChatGPT} to certain business tasks. Along with providing insights into how~\ac{ChatGPT}-powered expert systems may be effectively used in the business sector, the paper also identifies possible customer service application areas.\\
Businesses may increase customer loyalty and engagement by utilizing consumer data.~\ac{AI} applications have changed with the introduction of~\ac{ChatGPT}, although it requires modification to address certain business issues. In order to construct~\ac{ChatGPT}-powered expert systems for customer support, the authors in~\cite{sutrisno2023exploring} explores potential applications for their efficient use in the business sector using an iterative process that combines expert system development with prompt engineering.\\
According to~\cite{sudirjo2023application},~\ac{ChatGPT} can enhance comprehension of customer wants and satisfaction when used for consumer sentiment analysis.~\ac{ChatGPT} is useful for analyzing language and emotions, but for appropriate interpretation, it should be used in conjunction with human judgment. Important suggestions include properly organizing data, using appropriate datasets to train models, and using a variety of techniques to validate results.\\
\subsection{Robotics and Home Automation}
By offering natural language interfaces for device management, troubleshooting, and assistance,~\ac{ChatGPT} supports robotics and home automation. It facilitates smooth communication with automated devices, improving the user experience and increasing the intuitiveness of smart homes. In~\cite{wake2023chatgpt}, a few-shot approach of employing~\ac{ChatGPT} to translate natural-language commands into robot behaviors is presented. While resolving~\ac{ChatGPT}'s token limitations, customizable prompts allow for integration with robot systems, environment adaptation, and the creation of multi-step plans. Extensive record-keeping is eliminated by reusing environmental data in planning. Tests conducted in VirtualHome and residential settings demonstrate efficient work planning, with feedback leading to approximately 100\% accuracy and executability.\\
As~\ac{AI} advances, domestic service robots are becoming more prevalent in homes. The authors in~\cite{yang2024future} study the business potential and how~\ac{AI} improves the decision-making capabilities of robots such as CAESAR in domestic settings. It describes the functions of the three primary~\ac{AI} models—speech, visual, and language recognition—as well as how they are used in everyday life, education, and healthcare.\\
With the use of~\acp{LLM} like~\ac{GPT}-4, conversational agents like smart home assistants have the potential to advance sustainability. This study presents GreenIFTTT, a~\ac{GPT}-4-powered application for creating and controlling energy-efficient home automation routines. GreenIFTTT's effectiveness and usefulness were emphasized in an exploratory study conducted in December 2023 in Italy with 13 participants, showing that it has the potential to make home automation more ecologically friendly~\cite{giudici2024designing}.\\
The study~\cite{10368220} offers an open-source, reasonably priced framework for automating smart homes that integrates prototype and commercial sensors for improved interoperability. It prioritizes~\ac{AI}-driven solutions and future advancements for users, particularly those with limited mobility, by integrating an open-source~\ac{IoT} stack and the OpenAI API, which allows for intelligent, context-aware decision-making.
\subsection{Speech Recognition and Synthesis}
By translating spoken language into text and producing speech that sounds natural from text,~\ac{ChatGPT} facilitates speech recognition and synthesis. This facilitates more interactive voice-based applications and improves communication accessibility. The necessity for a conceptual framework to comprehend the many chatbot technologies and their potential in education, especially language learning, is addressed in the study~\cite{jeon2023beyond}. Three essential components of chatbot systems—goal-orientation, embodiment, and multimodality—are identified through an analysis of 37 studies. These components define eight different types of chatbots and twelve educational affordances.\\
To improve the training of Terabot, a dialogue system for psychiatric therapy, the authors in~\cite{gabor2023ai} investigate the use of~\ac{ChatGPT}. It is utilized to produce more training phrases, increasing the dataset by 112\%, in order to overcome the problem of having little real-life data for such a domain-specific system. Despite difficulties with speech recognition, testing using 2802 speech recordings from 32 patients revealed that~\ac{ChatGPT}-augmented data increased intent recognition accuracy by 13\%, reaching 86\% overall.\\
The authors in~\cite{poornima2024improving} use Transformers for text, CNNs for images, and LSTMs for voice to improve~\ac{ChatGPT}'s multimodal interactions. Their study highlights the significance of integrated modality processing for upcoming developments in customer service, education, and virtual assistants.\\
The study~\cite{kuzdeuov2024chatgpt} introduces an assistive mobile application that is used by blind and visually impaired people to engage in natural conversation with~\ac{ChatGPT}. The application offers a user-friendly interface for smooth interaction and makes use of speech recognition, text-to-speech, keyword detection, and voice activity detection.
\subsection{Text-based Games and Simulations}
By producing dynamic narratives, interactive dialogues, and decision-driven outcomes,~\ac{ChatGPT} improves text-based games and simulations. With just text, it produces immersive experiences that let players to interact with intricate events and rich storytelling.
The use of~\ac{ChatGPT} and other~\acp{LLM} as creative collaborators in game design is another~\ac{ChatGPT} application that is explained by the authors in~\cite{anjum2024ink}. The authors examine whether~\ac{AI} support enhances, detracts from, or offers a different feature to games created by humans. Three prototype games were made: one with a simple foundation, one with features contributed by humans, and one with~\ac{ChatGPT}-generated aspects. To assess the games' quality and preferences, a user study was carried out. In order to evaluate~\ac{AI}'s function in game design, the article examines player input and talks about how to convey creative intent to the~\ac{AI}.\\
Another application of~\ac{GPT} in games is presented in~\cite{yang2024gpt}. In this study, procedural content creation, mixed-initiative game design, mixed-initiative gameplay, playing games, and game user research are the five main uses of~\ac{GPT} in game research that are highlighted. With the goal of advancing game production and improving player experiences with cutting-edge~\ac{AI} technology, the evaluation identifies new trends and makes recommendations for future research areas in each field.\\
The capabilities of~\ac{ChatGPT} and~\ac{GPT}-4 in text-based games, where players engage with the game environment through dialogue, are examined in~\cite{tsai2023can}.~\ac{ChatGPT} performs competitively, however, it still has some small drawbacks, like the inability to infer game objectives, exploit prior information, or construct a world model.\\ 
The first~\ac{ChatGPT} Game Jam, which examined the application of~\acp{LLM} in game development and took place in May 2023, is the subject of~\cite{grow2023chatgpt}. OpenAI's~\ac{ChatGPT}, which allows users to build original games using text prompts, has drawn a lot of interest since its November 2022 release. In addition to offering insights on the process's present potential and constraints, the event sought to embrace and explore~\ac{LLM}-based game development.\\
The study~\cite{matthews2023academics} examines how 16 education scholars interpret papers generated by~\ac{ChatGPT} or humans using Turing's Imitation Game. The intricacy of~\ac{AI} technology was demonstrated by the fact that scholars could only recognize~\ac{AI}-generated texts 50\% of the time. Their choices were mostly impacted by voice, word choice, structure, job completion, and flow. The results show how important it is to develop ways to help teachers better handle~\ac{AI}'s position in the classroom.\\
As a result of~\ac{ChatGPT} contribution in game area, the research~\cite{taveekitworachai2023journey} investigates biases in story endings generated by~\ac{ChatGPT} for story-driven games. Despite instructions for neutral endings,~\ac{ChatGPT} consistently produces positive ones, reflecting potential societal biases or majority preferences.
\subsection{Text Classification and Moderation}
To tackle the issues, including a lack of training data, poor domain transferability, and the high cost of deploying large models, the study~\cite{zhao2023chatagri} investigates~\ac{ChatGPT}'s potential in agricultural text classification. Important elements like answer parsing and prompt creation are examined. The findings demonstrate that~\ac{ChatGPT} successfully addresses these problems, outperforming refined PLM-based techniques without the need for training data unique to agriculture. \\
The authors in~\cite{shi2023chatgraph} present a system that improves interpretability, hence augmenting~\ac{ChatGPT}'s text categorization capabilities with an interpretable linear classifier. It uses~\ac{ChatGPT} to extract structured information from raw data and display it as a knowledge graph. The method performs better for text categorization than using~\ac{ChatGPT} directly.\\
The Banking77 dataset is used in the study~\cite{loukas2023breaking} to examine the effectiveness of~\ac{GPT} models for few-shot text categorization in the financial industry. Using in-context learning with~\ac{GPT}-3.5 and~\ac{GPT}-4, it provides accurate results without requiring expensive GPU compute or technical know-how. Even with fewer instances, the results show that~\ac{GPT}-3.5 and~\ac{GPT}-4 perform better than fine-tuned non-generative models.\\
The authors in~\cite{soni2023comparing} use both automatic measures and human evaluations to assess ~\ac{ChatGPT}'s abstractive summarization ability. Humans find it difficult to distinguish between genuine and~\ac{ChatGPT}-generated summaries, but computer classifiers are able to do so.\\
With an emphasis on model parameters, prompt modifications, and repeated inputs, the study~\cite{reiss2023testing} investigates~\ac{ChatGPT}'s zero-shot text annotation and classification consistency. The results show that~\ac{ChatGPT}'s consistency falls short of scientific dependability criteria when it comes to the task of identifying website texts as news or not.\\
\subsubsection{Email Filtering}
By examining incoming messages and classifying them according to their content, urgency, or relevance,~\ac{ChatGPT} helps with email filtering. It enhances overall email management, lowers spam, and helps prioritize important emails. Using datasets in Chinese and English, the study~\cite{si2024evaluatingperformancechatgptspam} investigates~\ac{ChatGPT}'s capacity to detect spam emails through in-context learning. The study assesses the effects of changing the quantity of prompt demonstrations on performance.~\ac{ChatGPT} beats deep supervised learning models on the low-resource Chinese dataset but performs worse on a big English dataset when compared to benchmark techniques such as naive Bayes, SVM, LR, DNN, and BERT classifiers.\\
The study~\cite{de2024hey} identifies emails based on phishing patterns using~\ac{ChatGPT}'s machine learning and natural language processing capabilities, reaching a 98.4\% sensitivity rate and 75.75\% accuracy. Although it needs more work,~\ac{ChatGPT} exhibits promise when compared to more conventional rule-based techniques and programs like FortiSandbox. The study identifies areas for improvement to improve email security while highlighting AI's potential to mitigate phishing risks.\\
The efficacy of SpamAssassin, a well-known Bayesian spam filter, against spam emails altered by~\acp{LLM} such as~\ac{GPT}-3.5 Turbo is assessed in~\cite{josten2024investigating}. According to the findings, SpamAssassin identified a serious flaw by incorrectly classifying up to 73.7\% of~\ac{LLM}-modified spam emails as authentic. Simple dictionary-replacement attacks, on the other hand, only succeeded 0.4\% of the time. \\
\subsubsection{Content Moderation}
Although content moderation systems are crucial for maintaining user safety on online social networks, they frequently fall short in providing equitable attention to the requirements of minorities and vulnerable users; therefore, filtering inappropriate content is crucial. Furthermore, these systems have trouble with user-platform communication and customisation. The authors of~\cite{franco2023analyzing} suggest improving content moderation systems by including~\acp{LLM} into the enforcement pipeline in order to address these problems.\\
The study~\cite{li2024hot} investigates~\ac{ChatGPT}'s capacity to identify harmful content on social media by contrasting its output with human annotations for content that is Hateful, Offensive, and Toxic (HOT).~\ac{ChatGPT} classified non-HOT content consistently and with an accuracy of roughly 80\%. But "hateful" and "offensive" are sometimes grouped under "toxic." Although prompt choice affects~\ac{ChatGPT}'s performance, the results indicate that it can help with content control.\\
The study~\cite{aldahoul2024advancing} assesses~\acp{LLM} for content moderation, including Google Gemini, GPT, OpenAI's moderation model, and Meta Llama. The study demonstrates that~\acp{LLM} perform better than conventional techniques using datasets like tweets, reviews, and multimedia, providing greater accuracy and fewer errors, underscoring their potential for usage in social media and other platform content moderation.\\
The optimization of~\acp{LLM} for the implementation of private content moderation is studied in~\cite{ma2023adapting}. In content moderation, it contrasts generative and discriminative models and shows how reasoning can be used to minimize overfitting even when reasoning is not output explicitly during deployment. The study offers a comprehensive method for optimizing~\acp{LLM} for vertical domain applications, covering everything from data gathering to model training.\\
\subsection{Text Generation}
ChatGPT uses advanced language modeling to generate human-like text for applications ranging from automatic responses to creative writing, providing logical and contextually relevant information. In~\cite{yuan2023evaluatinggenerativemodelsgraphtotext}, the authors investigate the usage of~\acp{LLM} for zero-shot graph-to-text creation. Using two datasets, it compares the performance of~\ac{GPT}-3 and~\ac{ChatGPT} with refined models such as T5 and BART. The AGENDA and WebNLG datasets' respective BLEU ratings of 10.57 and 11.08 demonstrate the ability of generative models to generate language that is coherent and fluid. Error analysis, however, identifies problems with producing distorted or unnecessary information and comprehending semantic linkages. High macro-F1 scores are also obtained when BERT is employed to detect machine-generated text.\\
The study~\cite{lancaster2023artificial} explores how students may abuse~\ac{ChatGPT} and other similar systems for evaluations, endangering academic integrity. Although it investigates watermarking methods for generated text detection, it comes to the conclusion that they are not a complete answer. The study suggests areas for additional research and promotes cooperation between the educational community and~\ac{AI}.\\
The performance of~\acp{LLM} in the biomedical area is assessed in~\cite{JAHAN2024108189} using 26 datasets and 6 tasks. It indicates that~\acp{LLM} pre-trained on huge text corpora are specialized even in biomedical applications, as zero-shot~\acp{LLM} perform better on smaller datasets than fine-tuned models.\\
The study~\cite{10.1093/bioinformatics/btad557} assesses~\ac{ChatGPT}'s performance in biomedical text
 generation and mining issues such as question answering, relation extraction, and named entity recognition. With a BLURB score of 58.50 versus the state-of-the-art model's 84.30,~\ac{ChatGPT} demonstrated both its efficacy and limitations in the comprehension and creation of biomedical texts.\\
 The authors in [5] introduce a novel dataset, HPPT, which improves detectors for human-machine collaboration-generated texts, and proposes the "Polish Ratio" to measure~\ac{ChatGPT}'s modification to original human text.
\subsection{Emotion}
\ac{ChatGPT} assists with activities like sentiment analysis and sympathetic replies by analyzing and interpreting emotional clues in text. Accurately identifying complex emotions and eliminating biases in its evaluations are still difficult tasks, though.
Using the Dair-\ac{AI}/emotion dataset, the study~\cite{10330544} assesses~\ac{ChatGPT}'s~\ac{NLP} skills in emotion categorization, obtaining a 58\% accuracy rate. This demonstrates both its advantages and disadvantages for sentiment analysis jobs.\\
By examining emotional understanding, parallel emotional response, and empathetic personality, the study~\cite{10388208} assesses~\ac{ChatGPT}'s capacity for empathy. According to the results,~\ac{ChatGPT} can identify emotions 91.7\% of the time and respond with similar sentiments 70.7\% of the time. Although its empathy scores are lower than those of healthy people, they are higher than those of people with high-functioning autism or Asperger syndrome.\\
Through the analysis of voice and facial expressions in video streams, the study~\cite{10198928} investigates the integration of~\ac{AI} for mental health assessments. It draws attention to how~\ac{ChatGPT} and collaborative robots, or cobots, can improve communication with kids who have autism. Tested using the IEMOCAP database, a unique emotional recognition technique shows promise for use in the medical field.\\
By automating transcription, annotation, and augmentation, the study~\cite{10448130} explores the use of basic models, such as~\ac{ChatGPT}, to enhance Speech Emotion Recognition (SER) systems. The findings indicate that these models combine outputs from several~\acp{LLM} to improve annotation quality and improve SER performance through transcription. The study also emphasizes how annotating unlabeled speech samples can help to extend emotion datasets.
The authors in~\cite{10388198} explore the use of~\acp{LLM} for image emotion estimation by utilizing~\ac{GPT}-3.5 for inference and creating captions. Although accuracy varies by emotion, it is found that ~\ac{GPT}-3.5 can predict emotions from captions with reasonable accuracy. Potential applications of~\acp{LLM} in image-based emotion estimation are demonstrated by the study.\\
The authors in~\cite{10447102} introduce CLAP4Emo, a new framework based on contrastive language-audio pretraining for speech emotion retrieval utilizing natural language prompts. For datasets lacking training captions, the technique uses~\ac{ChatGPT} to produce emotion captions. According to experimental findings, CLAP4Emo preserves excellent precision while enhancing emotion retrieval diversity across five benchmark datasets.\\
The study~\cite{10557749} explores how a social robot can identify emotions in conversation using~\ac{ChatGPT}. By using sentiment analysis to compare human and robot evaluations of emotions, it discovers that both parties generally concur on the most common feeling. The potential of incorporating emotion recognition into robot interactions was demonstrated by the robot's altered mood assessment upon receiving emotion recognition data.\\
In order to improve emotion identification, the authors in~\cite{10731460} investigates the use of ~\ac{ChatGPT} to translate informal SNS texts into ordinary language. The study demonstrates that transformer models trained on~\ac{ChatGPT}-augmented emotion datasets perform better than those trained only on original texts. This method shows the promise of~\acp{LLM} for~\ac{NLP} data augmentation by improving emotion interpretation, especially in human-robot interactions.
\begin{table*}[ht]
\caption{List of General and Specific application of~\acp{LLM}/~\ac{ChatGPT}.}
\centering
\begin{tabular}{p{4.7cm}p{3.9cm}p{6.8cm}}
\hline
 \textbf{General Application} & \textbf{Specific Application} & \textbf{Applied paper}    \\
\hline
%azim1
Anomaly Detection & - & \cite{shao2022log,shao2022log,balasubramanian2023transformer,qi2023loggpt,mannam2023optimizing,lai2023intrusion,khediri2024enhancing,ouyang2023quality,wang2024visiongpt,gu2024anomalygpt}\\
\hline
%azim2
Anomaly Detection & Fake Detection & \cite{teo2024integrating,huang2023fake,wu2023cheap,koru2024detection,ayoobi2023looming}\\
\hline
%azim3
Conversation Development  & Chatbot & \cite{henno2023we,sudharson2023abstractive,raj2024role,alomar2024refactor,xiao2024devgpt,mohamed2024chatting,pereira2023fuzzy,eagalapati2024enriching,ukirde2023english,song2024innovative,selitskiy2024yet,liu2024transformer,deo2024analyzing}\\
\hline
%azim4
Healthcare & - & \cite{sathe2023comprehensive,raiaan2024review,shankar2024deep,kuzlu2023rise,rahman2023survey,alamchatgpt,mohammed2024chatgpt}\\
\hline
%azim5
Healthcare & Diagnostic Assistance & \cite{panagoulias2023evaluating,mosaiyebzadeh2023empowering}\\
\hline
%azim6
Healthcare & Patient Communication & \cite{dou2023shennonggpt,montagna2024llm,gebreab2024llm,malkiel2024segllm,natarajan2024enhancing,angel2023clinical}\\
\hline
%azim7
Code Generation and Completion & -& \cite{gong2024ast, li2022competition,lai2023ds,wang2022execution}\\
\hline
%azim8
Code Generation and Completion & Software Development& \cite{deo2024analyzing}\\
\hline
%azim9
Software Development & Content Creation & \cite{liu2022wanli}\\
\hline
%azim10
Content Creation & Social Media & \cite{Gokaslan2019OpenWeb, baumgartner2020pushshift, ayoobi2023looming}\\
\hline
%azim11
Education and Tutoring & - & \cite{raiaan2024review, dan2023educhat, alamchatgpt, angel2023clinical}\\
\hline
%azim12
Education and Tutoring & language Learning & \cite{radford2018improving, devlin2018bert, yang2019xlnet, mcnamee2004character, wang-etal-2018-glue, yang2019xlnet, lan2019albert, nguyen-etal-2024-culturax, xu-etal-2020-clue, xu-etal-2020-clue, wang2019superglue, yao2021cuge, saha2018duorc, nie-etal-2020-adversarial, yang-etal-2023-glue, hendrycks2020measuring, zhang-etal-2022-cblue}\\
\hline
%azim13
Education and Tutoring & Academic Research & \cite{zhang2024map, lo-etal-2020-s2orc}\\
\hline
%azim14
Education and Tutoring & Education and Assisting students & \cite{ibrahim2024does, ergene2024ai, MEMARIAN2023100022, 10442033, khoso2023use}\\
\hline
%azim15
Education and Tutoring & Writing Assistance & \cite{najah2024improving, hidayatullah2024evaluating, fitria2023artificial, sain2025benefits}\\
\hline
%azim16
Education and Tutoring & Research & \cite{ma2024iterative, yang2023exploring, zhang2023summit, tariq2023assessing, gao2023human}\\
\hline
%azim17
Language Modeling/Machine Translation & NLP: Improving translation& \cite{sennrich-etal-2016-improving, sennrich-etal-2016-neural, zhang2016google, gheini-etal-2021-cross, luong-etal-2015-effective, lewis2019bart, junczys-dowmunt-etal-2018-marian, zheng2024fine, przybocki2006edit, papineni2002bleu, koehn2005europarl, costa2022no}\\
\hline
%azim18
Customer Support & - & \cite{zierock2023leveraging, haleem2024exploring, limna2023role, ezenkwu2023towards, sutrisno2023exploring, sudirjo2023application}\\
\hline
%azim19
Automation & Robotics and Home Automation & \cite{wake2023chatgpt, yang2024future, giudici2024designing, 10368220}\\
\hline
%azim20
Speech Recognition  & - & \cite{jeon2023beyond, gabor2023ai, poornima2024improving, kuzdeuov2024chatgpt}\\
\hline
%azim21
Games & text-based Games  & \cite{anjum2024ink, yang2024gpt, tsai2023can, grow2023chatgpt, matthews2023academics, taveekitworachai2023journey}\\
\hline
%azim22
Text Classification  & - & \cite{zhao2023chatagri, shi2023chatgraph, loukas2023breaking, soni2023comparing, reiss2023testing}\\
\hline
%azim23
Text Classification & Email Filtering & \cite{si2024evaluatingperformancechatgptspam, de2024hey, josten2024investigating}\\
\hline
%azim24
Text Classification  & Filtering inappropriate content & \cite{franco2023analyzing, li2024hot, aldahoul2024advancing, ma2023adapting}\\
\hline
%azim25
Text Generation & - & \cite{yuan2023evaluatinggenerativemodelsgraphtotext, lancaster2023artificial, JAHAN2024108189, 10.1093/bioinformatics/btad557}\\
\hline
%azim26
Emotion & - & \cite{10330544, 10388208,10198928, 10448130, 10388198, 10447102, 10557749,10731460}\\
\hline
\end{tabular}
    \label{LLM_App}
\end{table*}

\section{Challenges and Ethical Considerations}\label{ethical}
\subsection{Bias and Fairness}
To ensure that big language models such as~\ac{ChatGPT} are reliable and fair, bias in training data and model outputs must be addressed. Prejudices from the past, preconceptions, and uneven portrayals of various groups can all contribute to biases in training data, which can then skew model outputs and reinforce these biases. Implementing strict data preprocessing methods that find and fix imbalances and include a variety of sample datasets are crucial for reducing this. In addition, biased behavior can be identified and corrected with the support of ongoing monitoring and assessment of model outputs. In order to identify and resolve potential biases, transparent documentation and interaction with a variety of stakeholders are also essential components of ethical~\ac{AI} development. We can work toward more fair and reliable~\ac{AI} systems by addressing these problems. \\
\subsection{Privacy Concerns}
Deploying big language models such as~\ac{ChatGPT} presents significant issues, two of which are data privacy and user security. Large volumes of data, sometimes containing sensitive personal information, are frequently needed for these models, which raises questions about data breaches and illegal access. To protect user information, strong encryption techniques and safe data storage procedures are crucial. Strict access controls and anonymization methods can also be used to further protect privacy. Retaining trust also depends on getting express user agreement and being transparent about data processing procedures. To ensure that user data is safeguarded during the model's lifecycle, ongoing assessments and modifications to security measures are required to handle new threats and vulnerabilities.\\
\subsection{Misinformation and Abuse} 
The usage of large language models such as~\ac{ChatGPT} raises serious concerns about misinformation and abuse. As a result of these algorithms' unintentional generation or amplification of incorrect information, disinformation may propagate. This risk is increased by their capacity to generate extremely logical and convincing writing, which is often abused by dishonest people for scams, propaganda, and other destructive endeavors. The problem is further complicated by the possibility that the models will be used to produce deepfakes or personas. The implementation of strict content moderation, the development of reliable detection methods for false information, and the promotion of digital literacy among users to enable them to critically assess~\ac{AI}-generated content are all necessary to mitigate these hazards. To successfully handle and manage these difficulties, it is imperative that researchers continue their work and collaborate with cybersecurity and ethics specialists. \\
\subsection{Regulatory and Policy Implications}
In order to control the implementation and effects of massive language models like~\ac{ChatGPT}, regulatory and policy consequences are essential. Clear rules on the moral application of these technologies must be established by policymakers to guarantee that they do not violate people's right to privacy or negatively impact society. Regulations ought to cover matters like data protection, material created by~\ac{AI} being held accountable, and the possibility of prejudice and discrimination in~\ac{AI} results. In order to keep an eye on and audit the application of these models, clear reporting and supervision procedures also need to be established. Governments, IT firms, and civil society organizations must work together to develop fair policies that promote innovation while defending the interests of the general public. Early and proactive regulatory frameworks can help reduce risks and guarantee the ethical adoption of~\ac{AI} technologies across a range of industries. \\ 
\section{Discussion and Conclusion}
\ac{AI}'s capacity to comprehend and produce content that is human-like has significantly improved as a result of the development of \acp{LLM} like \ac{ChatGPT}.  Applications in fields ranging from customer service and healthcare to education and security are made possible by \ac{ChatGPT}'s remarkable linguistic capabilities, which are a result of its transformer-based architecture, pre-training approaches, and fine-tuning procedures.  The broad dataset training and reinforcement learning techniques give \ac{ChatGPT} a degree of flexibility that improves its conversational and problem-solving skills.
\ac{ChatGPT} and other \acp{LLM} face many obstacles in spite of these benefits.  Algorithmic biases, privacy hazards, and the possibility of false information are examples of ethical issues that call for strict regulation and oversight.  Hallucinations, in which the model produces inaccurate or misleading data, continue to be a significant drawback that academics are currently attempting to address.  Model optimization approaches including parameter-efficient fine-tuning, quantization, and pruning are necessary to solve the efficiency and environmental issues raised by the significant computational resources needed to train and implement \acp{LLM}.
In contrast to existing \acp{LLM} like BERT, \ac{GPT}-4, and specialized domain-specific models, \ac{ChatGPT} performs exceptionally well in conversational tasks, while it might not always outperform in knowledge-intensive purposes.  For \ac{ChatGPT} and related systems to succeed in the future, continuous developments in multimodal models, federated learning for privacy-conscious \ac{AI}, and alignment strategies to improve moral \ac{AI} behavior will be essential.
In the future, research should concentrate on strengthening real-time adaptability, decreasing biases, improving \ac{LLM} transparency, and improving user interface frameworks.  For \acp{LLM} to be used responsibly, ethical standards and regulatory frameworks must change in tandem with technology.  With further development, \ac{ChatGPT} and its offspring could completely transform human-computer interactions and meaningfully close the gap between artificial and human intelligence.
%azimchange

%\input{acronyms1}

%\printglossaries % prints the acronym list
%\bibliographystyle{IEEEtran}
%\bibliography{references.bib}

\begin{acronym}[AAAAAAAAA]
    \acro{ABSA}{Aspect-Based Sentiment Analysis}
\acro{ACC}{Automatic Conversation}
\acro{AD}{autonomous driving}
\acro{AE}{autoencoding}
\acro{AI}{Artificial Intelligence}
\acro{ALBERT}{A Lite BERT}
\acro{API}{Application Programming Interface}
\acro{AR}{autoregressive}
\acro{APE}{Absolute Positional Encoding}
\acro{ASR}{Automatic Speech Recognition}
\acro{AST}{Abstract Syntax Tree}
\acro{BART}{Bidirectional and Auto-Regressive Transformer}
\acro{BERT}{Bidirectional Encoder Representations from Transformers}
\acro{BERF}{BERT Random Fores}
\acro{Biaffine}{Biaffine Attention for Dependency Parsing}
\acro{BiLSTM}{Bidirectional Long Short-Term Memory}
\acro{BioBERT}{Bidirectional Encoder Representations from Transformers for Biomedical Text Mining}
\acro{BLEU}{Bilingual Evaluation Understudy}
\acro{BLEURT}{Bilingual Evaluation Understudy with Representations from Transformer}
\acro{BPE}{Byte Pair Encoding}
\acro{CAP}{Contrastive pruning}
\acro{CBT}{Cognitive Behavioral Therapy}
\acro{ChatGPT}{Chat Generative Pre-trained Transformer}
\acro{CLIP}{Contrastive Language-Image Pre-training}
\acro{CLS}{Causal Language Modeling}
\acro{CNN}{Convolutional Neural Network}
\acro{COPA}{Choice of Plausible Alternatives}
\acro{CoT}{Chain of Thought}
\acro{CPU}{Central Processing Unit}
\acro{CR}{Comparative Ranking}
\acro{CRF}{Conditional Random Field}
\acro{CSR}{Conversational Speech Recognition}
\acro{CTC}{Connectionist Temporal Classification}
\acro{CXL}{Compute Express Link}
\acro{DEPN}{Detect and Editing Privacy Neurons}
\acro{DL}{Deep Learning}
\acro{DPO}{Direct Preference Optimization}
\acro{DRL}{deep reinforcement learning}
\acro{DRP}{Deep Reinforcement Learning}
\acro{DST}{Dialogue State Tracking}

\acro{ELMo}{Embeddings from Language Models}
\acro{E2E}{End-to-End}
\acro{EAE}{Event Argument Extraction}
\acro{ECQA}{Explanation and Commonsense Question Answering}
\acro{ELECTRA}{Efficiently Learning an Encoder that Classifies Token Replacements Accurately}
\acro{ERC}{Emotion Recognition in Conversation}
\acro{ESB}{Extended Summarization Benchmark}
\acro{FactScore}{Fact Consistency Scoring}
\acro{FL}{federated learning}
\acro{FLAN}{Fine-tuned Language Net}
\acro{FNN}{Feed-Forward Neural Network}
\acro{GDPR}{General Data Protection Regulation}
\acro{GPT}{Generative Pre-trained Transformer}
\acro{GPU}{Graphics Processing Unit}
\acro{GQA}{General Question Answering}
\acro{HF}{Hugging Face}
\acro{HIPAA}{Health Insurance Portability and Accountability Act}
\acro{IAD}{Industrial Anomaly Detection}
\acro{IDS}{Intrusion Detection System}
\acro{ILM}{Instruction Learning Model}
\acro{IoMT}{Internet of Medical Things}
\acro{IoT}{Internet of Things}
\acro{IoV}{Internet of Vehicles}
\acro{KD}{Knowledge Distillation}
\acro{KG}{Knowledge Graph}
\acro{LaCLIP}{Language Augmented CLIP}
\acro{LLaMA}{Large Language Model Meta AI}
\acro{LLM}{Large Language Model}
\acro{LLVM-AD}{Large Language and Vision Models for Autonomous Driving}
\acro{LM}{Language Model}
\acro{LPE}{Learnable Positional Embedding}
\acro{LSTM}{Long Short-Term Memory}
\acro{LVLM}{Large Vision-Language Model}
\acro{MAD}{Multiply-Add}
\acro{MAS}{multiagent systems}
\acro{MC}{Multiple Classification}
\acro{METEOR}{Metric for Evaluation of Translation with Explicit Ordering}
\acro{ML}{Machine Learning}
\acro{MLLM}{Multimodal Large Language Model}
\acro{MLM}{Masked Language Modeling}
\acro{MoE}{Mixture of Experts}
\acro{MS MARCO}{Microsoft MAchine Reading COmprehension}
\acro{MT}{Machine Translation}
\acro{NER}{Named Entity Recognition}
\acro{NLG}{Natural Language Generation}
\acro{NLI}{Natural Language Inference}
\acro{NLP}{Natural Language Processing}
\acro{NLU}{Natural Language Understanding}
\acro{NSFW}{Not Safe For Work}
\acro{OOV}{Out-Of-Vocabulary}
\acro{OPF}{Optimal Power Flow}
\acro{OTD}{One True Dialogue}
\acro{OWL}{Outlier Weighed Layer-wise}
\acro{PaLM}{Pathways Language Model}
\acro{PARA}{Paraphrase Identification}
\acro{PEFT}{Parameter Efficient Fine-Tuning}
\acro{PLO}{Posterior Log Odds}
\acro{PoS}{Part-of-Speech}
\acro{PPO}{Proximal Policy Optimization}
\acro{PTM}{Pre-trained Model}
\acro{PTQ}{Post-Training Quantization}
\acro{QA}{Question Answering}
\acro{QAQG}{Question Answering Question Generation}
\acro{QAT}{Quantization-Aware Training}
\acro{RAF}{Reasoning Accuracy Framework}
\acro{RLHF}{Reinforcement Learning from Human Feedback}
\acro{RLAIF}{Reinforcement Learning from AI Feedback}
\acro{RM}{Reward Modeling}
\acro{RoBERTa}{Robustly Optimized BERT Pre-training Approach}
\acro{RoPE}{Rotary Positional Embedding}
\acro{ROUGE}{Recall-Oriented Understudy for Gisting Evaluation}
\acro{RPE}{Relative Positional Encoding}
\acro{RNN}{Recursive Neural Network}
\acro{RSTB}{Residual Swin Transformer Block}
\acro{SA}{Surprise Adequacy}
\acro{SDS}{Spoken Dialogue System}
\acro{SFT}{Safety Fine-Tuning}
\acro{SIMD}{Single-Instruction Multiple-Data}
\acro{SLD}{Self-correcting Language-Driven}
\acro{SMOTE}{Synthetic Minority Over-sampling Technique}
\acro{SRAM}{Static Random Access Memory}
\acro{SST}{Sentiment Sentence Treebank}
\acro{STRN}{Swin Transformer and ResNet-based Generative Adversarial Network}
\acro{STU}{Source-Target Unification}
\acro{SVM}{Support Vector Machine}
\acro{SwinIR}{Swin Transformer-based image restoration}
\acro{T5}{Text-to-Text Transfer Transformer}
\acro{TAPEX}{Tabular Pretraining for Data Extraction and eXploration}
\acro{TF-IDF}{Term Frequency-Inverse Document Frequency}
\acro{THI}{Tinnitus Handicap Inventory}
\acro{TGI}{Textual Generative Intelligence}
\acro{TIP}{Topic-based Information Prioritization}
\acro{TPU}{Tensor Processing Unit}
\acro{TS}{Token Similarity}
\acro{TTS}{Text-to-Speech}
\acro{VQA}{Visual Question Answering}
\acro{WSD}{Word Sense Disambiguation}
\acro{XGLUE}{Cross-lingual General Language Understanding Evaluation}
\acro{GLUE}{General Language Understanding Evaluation}
\acro{SQuAD}{Stanford Question Answering Dataset}
\acro{RACE}{Reading Comprehension from Examinations}
\acro{DINO}{DIstillation with NO labels}
\acro{SCIBERT}{Scientific Bidirectional Encoder Representations from Transformers}
\acro{GP}{Genetic Programming}
%azim
\acro{XLNet}{Extra Long Transformer Network}

\acro{FFN}{Feed Frward Network}
\acro{RTD}{replaced token detection}
\acro{CLIP}{Contrastive Language–Image Pretraining}
\acro{SOTA}{scores of the state-of-the-art}
\acro{MRR}{Mean Reciprocal Rank}
\acro{ResNet}{Residual Network}
\acro{Swin}{Shifted Window}
\acro{COCO}{Common Objects in Context}
\acro{MT-NLG}{Megatron-Turing NLG}
\acro{AWS}{Amazon Web Services}
\acro{}{}
\acro{}{}
\acro{}{}
\acro{}{}
\acro{}{}
%afshin
\end{acronym}  % longest 

\end{document}